\pdfoutput=1

\documentclass[acmlarge]{acmart}

\usepackage{caption}
\usepackage{setspace}
\usepackage{enumitem}

\usepackage{subcaption}
\usepackage{hyperref}
\usepackage{bbm}
\usepackage{algorithm}
\usepackage{algorithmic}
\usepackage{tabularx}

\usepackage{diagbox}
\usepackage{flushend}
\usepackage{color}
\definecolor{cmtColor}{RGB}{0, 0, 0}

\def \eg {\emph{e.g.}, }
\def \ie {\emph{i.e.}, }
\def \th {\mathop{\mathrm{th}}}

\begin{document}

\title{RDeepSense: Reliable Deep Mobile Computing Models with Uncertainty Estimations}

\author{Shuochao Yao}
\affiliation{%
  \institution{University of Illinois Urbana Champaign}
  \streetaddress{201 North Goodwin Avenue}
  \city{Urbana}
  \state{IL}
  \postcode{61801}
  \country{USA}}
  
\author{Yiran Zhao}
\affiliation{%
  \institution{University of Illinois Urbana Champaign}
  \streetaddress{201 North Goodwin Avenue}
  \city{Urbana}
  \state{IL}
  \postcode{61801}
  \country{USA}} 
 
\author{Huajie Shao}
\affiliation{%
  \institution{University of Illinois Urbana Champaign}
  \streetaddress{201 North Goodwin Avenue}
  \city{Urbana}
  \state{IL}
  \postcode{61801}
  \country{USA}}
  
\author{Aston Zhang}
\affiliation{%
  \institution{University of Illinois Urbana Champaign}
  \streetaddress{201 North Goodwin Avenue}
  \city{Urbana}
  \state{IL}
  \postcode{61801}
  \country{USA}}
  
\author{Chao Zhang}
\affiliation{%
  \institution{University of Illinois Urbana Champaign}
  \streetaddress{201 North Goodwin Avenue}
  \city{Urbana}
  \state{IL}
  \postcode{61801}
  \country{USA}}
  
\author{Shen Li}
\affiliation{%
  \institution{IBM Research}
  \streetaddress{1101 Kitchawan Rd}
  \city{Yorktown Heights}
  \state{NY}
  \postcode{10598}
  \country{USA}}
  
\author{Tarek Abdelzaher}
\affiliation{%
  \institution{University of Illinois Urbana Champaign}
  \streetaddress{201 North Goodwin Avenue}
  \city{Urbana}
  \state{IL}
  \postcode{61801}
  \country{USA}}

\sloppy

\begin{abstract}
Recent advances in deep learning have led various applications to unprecedented achievements, which could potentially bring higher intelligence to a broad spectrum of mobile and ubiquitous applications. Although existing studies have demonstrated the effectiveness and feasibility of running deep neural network inference operations on mobile and embedded devices, they overlooked the reliability of mobile computing models. Reliability measurements such as predictive uncertainty estimations are key factors for improving the decision accuracy and user experience.
In this work, we propose RDeepSense, the first deep learning model that provides well-calibrated uncertainty estimations for resource-constrained mobile and embedded devices. 
RDeepSense enables the predictive uncertainty by adopting a tunable proper scoring rule as the training criterion and dropout as the implicit Bayesian approximation, which theoretically proves its correctness.
To reduce the computational complexity, RDeepSense employs efficient dropout and predictive distribution estimation instead of model ensemble or sampling-based method for inference operations.
We evaluate RDeepSense with four mobile sensing applications using Intel Edison devices. Results show that
RDeepSense can reduce around $90\%$ of the energy consumption while producing superior uncertainty estimations and preserving at least the same model accuracy compared with other state-of-the-art methods.
\end{abstract}

\maketitle

{

\section{Introduction}~\label{sec:introduction}
Using embedded sensors to infer the surrounding physical states and context is one of the major tasks of mobile and ubiquitous computing. Numerous mobile applications have prospered in a wide range of areas, such as health and wellbeing~\cite{wang2016hemaapp,kaiser2016design,bauer2012shuteye,faurholt2016behavioral,bentley2015reducing,griffiths2014health,toscos2012best}, behavior and activity recognition~\cite{choudhury2008mobile,chen2014airlink,melgarejo2014leveraging,yao2016learning,pu2013whole,mannini2013activity,yeo2016watchmi}, crowd sensing~\cite{yao2016recursive,yao2016source,zhang2017regions,zhang2017triovecevent}, tracking and localization~\cite{zhang2015eye,chung2011indoor,weppner2016monitoring,koehler2014indoor,grosse2016platypus,jiang2012ariel}. 
An important component in these applications is a learning model that outputs target values given sensor inputs. 
 
Rapid advancement in deep learning techniques has tempted researchers to employ deep neural networks as the learning models in the mobile applications. 
These highly capable models are good at making sophisticated mappings between unstructured data such as sensor inputs and target quantities, which can hardly be achieved by traditional machine learning models.
Specific deep learning models have been designed to fuse multiple sensory modalities and extract temporal relationships along sensor inputs. These specifically designed models have shown significant improvements on audio sensing~\cite{lane2015deepear}, tracking and localization~\cite{yao2017deepsense}, human activity recognition~\cite{radu2016towards, CastroHickson2015, yao2017deepsense, guan2017ensembles}, and user identification tasks~\cite{yao2017deepsense}.

However, the inability of treating the deep learning model more than just an incomprehensible black box has become an important factor that hinders researchers from applying the model to mobile applications. The complexity and uninterpretability of such models mainly result from the deep and non-linear structures~\cite{lipton2016mythos}. 
Therefore researchers can hardly understand how deep neural networks derive their final predictions. This leads to either the loss of trust in deep learning models or blind faith in deep learning models 
without being aware of predictive uncertainties and error bound. 

In order to explicitly output the reliability measure of deep neural network model, 
we aim to provide the model with predictive uncertainties during inference.
Predictive uncertainty is defined as the probability of occurrence of the target variable conditioned on all available information. 
One particular approach to express predictive uncertainty is to treat the model predictions as random variables, \ie in the form of probability distributions instead of point estimations~\cite{quinonero2006evaluating}. In this paper, we center our discussion around this specific representation of the uncertainty.

On one hand, although it is hard to directly interpret deep neural networks, predictive uncertainty can provide the quantitative confidence of prediction correctness, 
which boosts trust and faith in deep learning models.
On the other hand, uncertainty estimation itself is crucial for 
scientific measurements~\cite{krzywinski2013points, ghahramani2015probabilistic}. 
Extensive investigations show that measurement uncertainties can impact user experiences~\cite{lim2011investigating,kay2015good}.
In order to monitor the uncertainties of mobile sensing applications, 
the first important step is to obtain the predictive uncertainties of learning models used in the applications, which, in our case, are deep neural network models.

However, enabling deep neural networks to provide high-quality and well-calibrated uncertainty estimations on mobile and embedded devices poses two major challenges. One challenge is to provide a mathematically grounded uncertainty estimations that require few changes on either the model or the optimization method.
Although mathematically grounded methods such as Bayesian approaches serves as powerful tools to estimate predictive uncertainties~\cite{clyde2004model},
Bayesian neural networks are computationally expensive to train and inference 
even for brawny servers, let alone mobile and embedded devices~\cite{quinonero2006evaluating}. 
Therefore a mathematically grounded theory under minimal model modification requirements is a must for reliable uncertainty estimations.

The other challenge is to reduce the computational burden of uncertainty estimations during inference. For mobile and ubiquitous computing applications,  although we can train the deep neural networks on brawny servers with powerful GPUs, 
running inference on mobile and embedded devices is difficult due to limited energy supplies and computational resources on such devices~\cite{yao2017deepiot}.
Illuminating studies from the machine learning community try to provide mathematically grounded uncertainty estimations for deep neural networks, but these methods are based either on the sampling method~\cite{Gal2016DropoutAA} or the ensemble method~\cite{lakshminarayanan2016simple}. They require either running a single stochastic neural network for multiple times or training and running multiple deterministic neural networks. All these solutions are not resource-friendly to mobile and embedded devices. 
Therefore, mobile applications call for a novel solution that theoretically guarantees the correctness of predictive uncertainties, and at the same time consumes much less resource. 

In this work, we propose RDeepSense that enables predictive uncertainties with theoretically proven correctness for mobile and ubiquitous applications. RDeepSense significantly reduces the computational overhead and preserves at least the same model accuracy. To the best of our knowledge, this is the first deep learning model that provides uncertainty estimations for resource-limited devices.
The core of RDeepSense is the integration of the dropout training method that interprets neural networks as Gaussian process (GP) through Bayesian approximation~\cite{srivastava2014dropout,rasmussen2006Gaussian,Gal2016DropoutAA} and proper scoring rules as training criterion that measure the quality of predictive uncertainty such as log-likelihood and the Brier score ~\cite{gneiting2007strictly}. Their integration can be further interpreted as the mixture distribution of a Gaussian or categorical distribution based on latent deep Gaussian process and a deep Gaussian process
 through Bayesian approximation. Firstly, RDeepSense uses a tunable proper scoring rule as the training criterion that significantly mitigates the problem of underestimating predictive uncertainties in deep neural networks~\cite{Gal2016DropoutAA}. Secondly, since dropout training can be interpreted as ``geometric averaging" over the ensemble of possible ``thinned" subnetworks~\cite{baldi2013understanding}, RDeepSense applies dropout training instead of model ensemble.
It greatly reduces the computation complexity of the final neural network compared with model ensemble. 
Therefore, our integrated method incurs only little computational overhead, which makes it feasible on embedded devices for mobile applications.

Evaluations of RDeepSense use
the Intel Edison computing platform~\cite{Edison}. 
We conduct mobile and ubiquitous tasks that focus on human health and wellbeing, smart city transportation, environment monitoring, and activity recognition. 
Specifically, our experiments include: 1) monitoring arterial blood pressure through photoplethysmogram (PPG) from fingertip~\cite{kachuee2015cuff}, 2) NY city taxi commute time estimation~\cite{wang2016simple}, 3) gas mixture concentrations estimation through the chemical sensor array~\cite{fonollosa2015reservoir}, 4) and heterogeneous human activity recognition through motion sensors~\cite{stisen2015smart}.

We compare RDeepSense with the state-of-the-art Monte Carlo dropout method~\cite{Gal2016DropoutAA}, ensemble method~\cite{lakshminarayanan2016simple}, and Gaussian process.
The resource consumption of Intel Edison module and final model performance such as  the accuracy and the quality of uncertainty estimations are measured for all the algorithms. RDeepSense can reduce more than $90\%$ of inference time and energy consumption, while obtaining the uncertainty estimations with better quality compared with the other algorithms. 
The 
well-calibrated
uncertainty estimations and resource efficiency make RDeepSense the first choice to obtain uncertainty estimations of deep neural networks in mobile applications.

In summary, we propose a simple yet effective and theoretically-grounded method, RDeepSense, which empowers neural networks with 
well-calibrated
predictive uncertainty estimations. RDeepSense is also a resource-friendly algorithm for mobile and embedded devices that adds almost no computational overhead during model inference.

The rest of paper is organized as follows. Section~\ref{sec:related} introduces related works about uncertainty estimations and deep neural networks. We describe the technical details of RDeepSense in Section~\ref{sec:RDeepSense}. The evaluation is presented in Section~\ref{sec:evaluation}. Finally,
we discuss the results in Section~\ref{sec:discussion} and conclude in Section~\ref{sec:conclusion}.

\section{Related work}~\label{sec:related}
On one hand, reliability and uncertainty estimation is one important issue of mobile and ubiquitous computing. A lot of works have been proposed to utilize uncertainty estimations for improving the decision accuracy and user experience. Baumann et al.~\cite{baumann2016quantifying} make next-place predictions based on the uncertainty estimation of classifiers. Kay et al.~\cite{kay2016ish} propose a novel discrete representation of uncertainties for visualizing and user interaction. Boukhelifa et al.~\cite{boukhelifa2017data} propose design considerations for uncertainty-aware data
analytics. On the other hand, the recent advances in deep learning techniques have motivated people to apply deep neural networks for solving mobile and ubiquitous computing tasks. Lane et al.~\cite{lane2015deepear} apply deep neural networks to solve audio sensing tasks. Castro et al.~\cite{CastroHickson2015} predict daily activities from egocentric images using deep learning. Yao et al.~\cite{yao2017deepsense} propose a deep learning structure that fuses multiple sensor inputs and extracts time dependencies. 
{\color{cmtColor}Guan et al.~\cite{guan2017ensembles} apply ensembles of LSTM for activity recognition.}
However, uncertainty estimations of deep neural networks for mobile and ubiquitous computing tasks is an important topic that draws less attention.

\begin{table}[!htb]
%\vspace{-0.275cm}
\footnotesize
\begin{center}
\caption {Comparison among deep learning based predictive uncertainty estimation}
\vspace{-0.3cm}
\hspace{-0.1cm}
\label{tab:algorithm}
\begin{tabular}{ |c | c | c | c | c | c | c | } 
 \hline
 Algorithm & Dropout Training & Proper Scoring Rules  &  Ensemble method & Obtain predictive uncertainty with single run  \\ 
  \hline
   \hline
  RDeepSense & $\checkmark$ & $\checkmark$ & $\times$& $\checkmark$   \\ 
  \hline
  MCDrop &$\checkmark$ & $\times$& $\times$ &  $\times$ \\ 
 \hline
 SSP & $\times$& $\checkmark$  & $\checkmark$ & $\times$  \\ 
 \hline
\end{tabular}
\end{center}
%\vspace{-0.425cm}
\end{table} 

Recently there are some illuminating works from the machine learning community that tries to provide deep neural networks with uncertainty estimations. Gal et al.~\cite{Gal2016DropoutAA} provide the first theoretical proof of the linkage between dropout training with deep Gaussian process called MCDrop. However, the proposed method tends to underestimate the uncertainty due to the nature of variational inference. Lakshminarayanan et al.~\cite{lakshminarayanan2016simple} propose a solution SSP based on proper scoring rules and ensemble methods. However, the proposed method tends to overestimate the uncertainty on real datasets.

Since these previous works do not consider the scenario of mobile and ubiquitous computing, all these proposed methods require the operations with high computational cost during model inference, \ie sampling methods or ensemble methods. These computationally intensive operations aggravate the time and energy consumption problems in the embedded devices, which is one of the key issues of mobile and ubiquitous computing~\cite{park2011gesture,pirkl2012robust,gordon2012energy}.

To the best of our knowledge, RDeepSense is the first work that provides a simple yet effective solution to estimate the uncertainties of deep neural networks for mobile and ubiquitous computing applications. RDeepSense uses proper scoring rules to mitigate the underestimation effect of MCDrop, and applies dropout training as implicit ensemble to avoid the computationally intensive ensemble method used in SSP.

In order to further illustrate the main difference between RDeepSense and other two deep leaning uncertainty estimation algorithms, MCDrop and SSP, we show the designing components of these three algorithms in Table~\ref{tab:algorithm}.

\section{RDeepSense Framework}~\label{sec:RDeepSense}
This section elaborates on the technical details
of the RDeepSense framework in three constituents. 
Section~\ref{sec:RDeepSense_basic} introduces 
a simple yet effective recipe to build a fully-connected neural network
with predictive uncertainty estimations.
In Section~\ref{sec:analysis}, we introduce preliminary knowledge and make the theoretical analysis of RDeepSense. We prove that RDeepSense is a mathematically grounded method to obtain predictive uncertainty estimations. In Section~\ref{sec:RDeepSens_run}, we introduce an effective and efficient approximation for RDeepSense to obtain predictive uncertainty estimations while running on the resource-constrained embedded devices.

For the rest of this paper, all vectors are denoted by bold lower-case letters (\eg $\mathbf{x}$ and $\mathbf{y}$), and matrices and tensors are represented by bold upper-case letters (\eg $\mathbf{X}$ and $\mathbf{Y}$). For a column vector $\mathbf{x}$, the $j^{\th}$ element is denoted by $\mathbf{x}_{[j]}$. For a tensor $\mathbf{X}$, the $t^{\th}$ matrix along the third axis is denoted by $\mathbf{X}_{\cdot \cdot t}$, and the other slicing notations are defined similarly. 
The superscript $l$ in $\mathbf{x}^{(l)}$ and $\mathbf{X}^{(l)}$ 
denote the vector and tensor for the $l^{\th}$ layer of the neural network. We use calligraphic letters to denote sets (\eg $\mathcal{X}$ and $\mathcal{Y}$), where $|\mathcal{X}|$ denotes the cardinality of $\mathcal{X}$.

\subsection{RDeepSense components}~\label{sec:RDeepSense_basic}
RDeepSense is a simple and effective method that empowers fully-connected neural networks to output predictive uncertainty estimations. There are only two steps to convert an arbitrary fully-connected neural networks into a neural network with uncertainty estimations:

\begin{enumerate}
\item Insert dropout operation to each fully-connected layer.~\label{enu:dropout}
\item Adopt a proper scoring rule as the loss function, and emit a distribution estimation instead of a point estimation at the output layer.~\label{enu:distout}
\end{enumerate}

The following two subsections describe dropout training and proper scoring rules in detail.

\subsubsection{Dropout training}~\label{sec:dropout}
Fully-connected neural networks can be formulated using the following equations:
\begin{equation}
\begin{split}
\mathbf{y}^{(l)} &= \mathbf{x}^{(l)} \mathbf{W}^{(l)}  + \mathbf{b}^{(l)},\\
\mathbf{x}^{(l+1)} &= f^{(l)}\big(\mathbf{y}^{(l)}\big),
\end{split}
\label{eqn:fcnet}
\end{equation}
where the notation $l = 1, \cdots, L$ is the layer index in the fully-connected neural network. For any layer $l$, the weight matrix is denoted as $\mathbf{W}^{(l)}\in \mathbb{R}^{d^{(l-1)}\times d^{(l)}}$; the bias vector is denoted as $\mathbf{b}^{(l)}\in \mathbb{R}^{d^{(l)}}$; the input is denoted as $\mathbf{x}^{(l)} \in \mathbb{R}^{d^{(l-1)}}$; and $d^{(l)}$ is the dimension of the $l^{\th}$ layer.  In addition, $f^{(l)}(\cdot)$ is a nonlinear activation function. 

However, such formulations could run into feature co-adapting and model overfitting problems. To avoid these problems, researchers introduce the concept of dropout as a regularization method~\cite{srivastava2014dropout}. ``Dropout" originally refers to dropping out hidden and visible units in a neural network, which is mathematically equivalent to ignoring rows of the weight matrix $\mathbf{W}^{(l)}$. Therefore, a fully-connected neural network with dropout can be represented as follows:
\begin{equation}
\begin{split}
\mathbf{z}_{[i]}^{(l)} &\sim  \text{Bernoulli}(\mathbf{p}_{[i]}^{(l)}), \\
\tilde{\mathbf{W}}^{(l)} &=  \text{diag}\big( \mathbf{z}^{(l)}\big) \mathbf{W}^{(l)}, \\
\mathbf{y}^{(l)} &= \mathbf{x}^{(l)} \tilde{\mathbf{W}}^{(l)}  + \mathbf{b}^{(l)},\\
\mathbf{x}^{(l+1)} &= f^{(l)}\big(\mathbf{y}^{(l)}\big).
\end{split}
\label{eqn:dropout}
\end{equation}

As shown in \eqref{eqn:dropout}, a vector of Bernoulli variables $\mathbf{z}^{(l)} \in \{0,1\}^{d^{(l-1)}}$ 
forms a diagonal matrix which acts as a mask to dropout the $i^{\th}$ row of $\tilde{\mathbf{W}}^{(l)}$ with probability $\mathbf{p}_{[i]}^{(l)}$. 
Intuitively, the dropout operations \eqref{eqn:dropout} convert a traditional (deterministic) neural network with parameters $\{\mathbf{W}^{(l)}\}$ into a random Bayesian neural network with random variables $\{\tilde{\mathbf{W}}^{(l)}\}$, 
which equates a neural network with a statistical model without using the Bayesian approach explicitly. 
This conversion with dropout helps us to obtain predictive uncertainty estimations and avoid the computationally intensive operations used in Bayesian approaches. The detailed analysis about the equivalence will be discussed later.

\subsubsection{Proper scoring rules}~\label{sec:scoring}
Optimizing a deep neural network requires minimizing the loss function. Therefore the loss function plays a crucial role in designing an effective neural network.
Many commonly used neural network loss functions are proper scoring rules, such as logistic loss and hinge loss. 

Scoring rules, also known as score functions, measure the quality of predictive uncertainties~\cite{gneiting2007strictly}.
Assume that $p_{\theta}(y|\mathbf{x})$ is the probabilistic distribution represented by a deep neural network. The scoring rule $S(p_{\theta}(y|\mathbf{x}), (\mathbf{x}, y))$ assigns a numerical score for the quality of predictive distribution $p_{\theta}(y|\mathbf{x})$ on event $(\mathbf{x}, y)\sim q(\mathbf{x}, y)$, where $q(\mathbf{x}, y)$ is the true distribution of data samples. 
The expected scoring rule is formulated as
\begin{equation}
S(p_{\theta}(y|\mathbf{x}), q(\mathbf{x}, y)) = \int q(\mathbf{x}, y) S(p_{\theta}(y|\mathbf{x}), (\mathbf{x}, y)) d\mathbf{x} dy \label{eqn:expected_score}.
\end{equation}

For a proper scoring rule, the equality in $S(p_{\theta}(y|\mathbf{x}), q(\mathbf{x}, y)) \ge S(q(\mathbf{x}, y)), q(\mathbf{x}, y))$ holds if and only if $p_{\theta}(y|\mathbf{x}) = q(\mathbf{x}, y)$.  Widely-adopted proper scoring rules include Log-likelihood $\log p_{\theta}(y|\mathbf{x})$ and Brier score $-\sum_{k=1}^K(\mathbbm{1}_k(y) -p_{\theta}(y=k|\mathbf{x}))^2$.

RDeepSense employs a tunable function, the weighted sum of negative log-likelihood and mean square error (Brier score for classification problems), which is a proper scoring rule, as the loss functions for both regression and classification problems. This loss function tries to offset the effect of overestimation and underestimation caused by negative log-likelihood and mean square error respectively, which will be analyzed and evaluated later.

For regression problems, in order to optimize the neural network with negative log-likelihood, we have to emit a distribution estimation instead of a point estimation at the output layer. Therefore, we slightly change the structures of neural networks. {\color{cmtColor}The last output layer generates both the predictive mean $\mu(\hat{y})$ and the predictive variance $\sigma^2(\hat{y})$.}
According to the notation in \eqref{eqn:dropout}, the output layer is represented by
$\mathbf{x}^{L+1} = \big[\mu(\hat{y}),\; \sigma^2(\hat{y})\big]^\intercal = \big[\mathbf{y}_{[0]}^{(L)},\; \mathop{\mathrm{softplus}}(\mathbf{y}_{[1]}^{(L)})\big]^\intercal$, where softplus function is $\log(1+\exp(\cdot))$ enforcing the positivity constraint on the variance. Predictive mean $\mu(\hat{y})$ and predictive variance $\sigma^2(\hat{y})$ compose a Gaussian distribution $\mathcal{N}(\mu(\hat{y}), \sigma^2(\hat{y}))$ as the output predictive distribution of the neural network.

Then the final loss function of a regression problem, $\mathcal{L}_{r}$, is the weighted sum of mean square error $\mathcal{L}_{re}$ and negative log-likelihood $\mathcal{L}_{rl}$,
\begin{equation}
\begin{split}
\mathcal{L}_{re} &= \sum_{n=1}^N \big(y - \mu(\hat{y})\big)^2 + \lambda_e\sum_{l=1}^L \|\mathbf{W}^{(l)}\|_2^2,\\
\mathcal{L}_{rl} &= \sum_{n=1}^N\Big(\frac{1}{2}\log \sigma^2(\hat{y}) + \frac{1}{2 \sigma^2(\hat{y})} \big(y - \mu(\hat{y})\big)^2\Big) + \lambda_l\sum_{l=1}^L \|\mathbf{W}^{(l)}\|_2^2, \\
\mathcal{L}_{r} &= (1-\alpha) \cdot \mathcal{L}_{rl} + \alpha \cdot \mathcal{L}_{re},
\end{split}
\label{equ:loss_regression}
\end{equation}
where $N$ is the number of training samples, the second term in the first two equations are the $L_2$ regularization, and $\alpha$ is a hyper-parameter. 

{\color{cmtColor}As we will discuss in Section~\ref{sec:weighted} and evaluate in Section~\ref{sec:alpha_evaluation}, a larger $\alpha$ leads neural networks to focus more on estimating an accurate mean value, which may underestimate the true uncertainties, while a smaller $\alpha$ leads neural networks to estimate a larger variance during the optimization process, which may overestimate the true uncertainties. Therefore, $\alpha$ is a hyper-parameter that makes the bias-variance tradeoff and is tuned to generate a well-calibrated predictive uncertainty, \ie neither underestimation nor overestimation.}

For the classification problem, $f^{(L)}(\cdot)$ is the softmax function that generates predictive probabilities for each category. 
The final loss function of a classification problem, $\mathcal{L}_{c}$, is the weighted sum of mean square error $\mathcal{L}_{ce}$ and negative log-likelihood $\mathcal{L}_{cl}$,
\begin{equation}
\begin{split}
\mathcal{L}_{ce} &= \sum_{n=1}^N\sum_{k=1}^K(\mathbbm{1}_k(y) -p_{\theta}(y=k|\mathbf{x}))^2 + \lambda_e\sum_{l=1}^L \|\mathbf{W}^{(l)}\|_2^2,\\
\mathcal{L}_{cl} &= \sum_{n=1}^N-\log p_{\mathcal{W}}(\hat{y}=y|\mathbf{x}) + \lambda_l\sum_{l=1}^L \|\mathbf{W}^{(l)}\|_2^2, \\
\mathcal{L}_{c} &= (1-\alpha) \cdot \mathcal{L}_{cl} + \alpha \cdot \mathcal{L}_{ce},
\end{split}
\label{equ:loss_classification}
\end{equation}
where $N$ is the number of training samples, $K$ is the number of classes, the second term in the first two equations are the $L_2$ regularization, and $\alpha$ is a hyper-parameter.

In summary, the whole neural network is optimized through a tunable proper scoring rule that maximizes the quality of predictive uncertainties. The detailed theoretical backup and proof of the equivalence between RDeepSense and a statistical model will be shown in Section~\ref{sec:analysis}.

\subsection{Theoretical analysis: the equivalence between RDeepSense and statistical models}~\label{sec:analysis}
Uncertainty estimations are usually inferred by a statistical model, such as a gaussian process~\cite{rasmussen2006Gaussian} and a graphical model~\cite{koller2009probabilistic}.
This section provides the theoretical bases for using RDeepSense to estimate predictive uncertainties by proving the equivalence between the RDeepSense model and a statistical model. To achieve this goal, {\color{cmtColor}we first summarize the preliminary knowledge about the equivalence between dropout training with mean square error and a deep Gaussian process, which is proposed by Gal et al.~\cite{Gal2016DropoutAA} in Section~\ref{sec:gp}.} Then we prove the equivalence between dropout with the proper scoring rule (log-likelihood) and a Gaussian or categorical distributions based on latent deep Gaussian process in Section~\ref{sec:intergration}. Finally, in Section~\ref{sec:weighted}, we generalize the analysis to another tunable proper scoring rule, weighted sum of log-likelihood and negative mean square error, which provides the theoretical foundation for the RDeepSense.

{\color{cmtColor}
\subsubsection{Preliminary: Dropout with mean square error}~\label{sec:gp}}
Gaussian process is a powerful statistical tool that allows us to model distribution over functions~\cite{rasmussen2006Gaussian}. {\color{cmtColor}Here we introduce the preliminary knowledge about Gaussian process 
and its basic relationship with dropout training through variational approximation, which is first discussed and proven by Gal et al.~\cite{Gal2016DropoutAA}.}

Assume that we have $N$ pairs of training data, which can be formed into the input matrix $\mathbf{X} \in \mathbb{R}^{N\times d^{(0)}}$ and the corresponding output matrix $\mathbf{Y}\in \mathbb{R}^{N\times d^{(L)}}$. For the regression problem, we place a joint Gaussian distribution over all function values
\begin{equation}
\begin{split}
p(\mathbf{F} | \mathbf{X}) &\sim \mathcal{N}(0,K(\mathbf{X},\mathbf{X})),\\
p(\mathbf{Y} | \mathbf{F}) &\sim \mathcal{N}(F,\tau^{-1}\mathbf{I}).
\end{split}
\label{eqn:gp_org}
\end{equation}
where $\tau$ is the precision hyper-parameter and $K(\cdot,\cdot)$ is the covariance function, encoding the prior function distribution of the Gaussian process. With a dataset of $N$ samples, $K(\cdot, \cdot)$ is a $N\times N$ matrix.

To formulate a fully-connected neural network as a Gaussian process, for a single fully-connected layer in a Bayesian neural network, we can define the covariance function as 
\begin{equation}
K(\mathbf{x}, \mathbf{x}') = \int p(\mathbf{W}^{(l)})f^{(l)}(\mathbf{x}\mathbf{W}^{(l)} + \mathbf{b}^{(l)})f^{(l)}(\mathbf{x}'\mathbf{W}^{(l)} + \mathbf{b}^{(l)}) d\mathbf{W}^{(l)},
\label{eqn:gp_kernel}
\end{equation}
where $p(\mathbf{W}^{(l)})=\mathcal{N}(0, l^{-2}\mathbf{I})$ and $f^{(l)}(\cdot)$ is the nonlinear activation function.
For an $L$-layer fully-connected neural network, we can feed the output of one Gaussian process to the covariance of the next as a deep Gaussian process model~\cite{damianou2013deep}.
Then our final target, predictive distribution estimation, can be formulated as
\begin{equation}
p(\mathbf{y}|\mathbf{x}, \mathbf{X}, \mathbf{Y}) = \int p(\mathbf{y}|\mathbf{x}, \mathcal{W}) p(\mathcal{W}|\mathbf{X}, \mathbf{Y}) d\mathcal{W},
\label{eqn:gp_predict}
\end{equation}
where $p(\mathbf{y}|\mathbf{x}, \mathcal{W})$ is the whole Bayesian neural network with random variables $\mathcal{W} = \{p(\mathbf{W}^{(l)})\}$. 

However, calculating the predictive distribution estimation $p(\mathbf{y}|\mathbf{x}, \mathbf{X}, \mathbf{Y}) $ requires the posterior distribution $p(\mathcal{W}|\mathbf{X}, \mathbf{Y})$, and calculating the posterior distribution $p(\mathcal{W}|\mathbf{X}, \mathbf{Y})$ further requires calculating the inverse of an $N\times N$ matrix, which is infeasible for a large-scale dataset used by a deep neural network. Therefore, a variational distribution $q(\mathcal{W})  = \prod_{l=1}^L p(\tilde{\mathbf{W}}^{(l)})$ is proposed to approximate the true posterior distribution, where $\tilde{\mathbf{W}}^{(l)}$ is the random variable used in dropout operations introduced in \eqref{eqn:dropout}.

Then we minimize the KL divergence between the approximated posterior $q(\mathcal{W})$ and the posterior of the deep Gaussian process over the variational parameters $\{\tilde{\mathbf{W}}^{(l)}\}$. The minimization objective is the negative log evidence lower bound derived from the likelihood,
\begin{equation}
\mathcal{L}_{gp} = -\int q(\mathcal{W})\log p(\mathbf{Y}|\mathbf{X}, \mathcal{W}) d\mathcal{W} + KL(q(\mathcal{W})| p(\mathcal{W})).
\label{eqn:gp_kl}
\end{equation}

We can use Monte Carlo sampling to approximate the first integral in \eqref{eqn:gp_kl}, and \eqref{eqn:gp_kl} can be reduced to
\begin{equation}
\mathcal{L}_{gp} = \sum_{n=1}^N(\mathbf{y}_n - \hat{\mathbf{y}}_n)^2 +\frac{p_i l^2}{2\tau N}\sum_{l=1}^L \|\mathbf{W}^{(l)}\|_2^2,
\label{eqn:gp_vi}
\end{equation}  
where $p_i$ is the dropout probability in \eqref{eqn:dropout}, $\tau$ is the hyperparameter in \eqref{eqn:gp_org}, and $l$ is the length-scale used to define the prior distribution $p(\mathbf{W}^{(l)})$.

If we compare \eqref{eqn:gp_vi} with the first equations in \eqref{equ:loss_regression} and \eqref{equ:loss_classification}, we can find that optimizing a variational approximation of deep Gaussian process is equivalent to optimizing an dropout neural network based on mean square error as the loss function. 

However, mean square error is not a proper scoring rule for regression problems, which cannot generate a well calibrated uncertainty estimations. Besides, due to the mode matching nature of KL divergence, the variational approximating usually generates a highly underestimated predictive uncertainty~\cite{blei2017variational}, which is also verified in our experiments in Section~\ref{sec:accuracy_uncertainty}. Therefore we further discuss the case of dropout training with proper scoring rules in Section~\ref{sec:intergration} and Section~\ref{sec:weighted}, which enables RDeepSense to provide a high quality uncertainty estimation. 

\subsubsection{Dropout with negative log-likelihood}~\label{sec:intergration}
We have introduced the previous work that treats a neural network with dropout training based on mean square error loss function as a deep Gaussian process with variational approximation. We call this method MCDrop.

However, there are two drawbacks for MCDrop.
{\color{cmtColor}
One is the underestimation of predictive distribution. Variational Bayesian used in MCDrop is known to provide underestimated posterior uncertainty, because optimizing the KL divergence will generate a low-variance estimation to a single mode of true posterior distribution~\cite{blei2017variational}. In addition, the loss function of MCDrop is not a proper scoring rule that can help to mitigate the negative effect of underestimation caused by the variational Bayesian method. Underestimation is not a desirable property for mobile and ubiquitous computing applications, because it means that the deep neural network will always be over-confident about its prediction results.}

The other drawback of MCDrop is the high computational burden during uncertainty estimation. Since the output of MCDrop is a stochastic point estimation, Monte Carlo sampling method is required to estimate the predictive mean and variance. Therefore we need to run the whole neural network for multiple times, \ie running $k$ times for $k$ samples, to generate the predictive uncertainty. Since running time and energy consumption are two crucial problems for mobile and ubiquitous computing applications, MCDrop is not a suitable solution for applications running on embedded devices.

Therefore, we integrate proper scoring rules and dropout training in RDeepSense to solve the aforementioned two drawbacks. The proper scoring rules such as log-likelihood help to reduce or even erase the underestimation effect of MCDrop, because proper scoring rule is a score function that gives higher quality uncertainty estimations more credits. In addition, since a neural network with proper score rule directly generates a predictive distribution estimation instead of a point estimation, we can efficiently obtain an approximated expectation of uncertainty estimation through dropout inference. At the same time, dropout as Bayesian approximation can provide a equivalence between the deep neural network and a statistical model, which guarantees RDeepSense to be a mathematically grounded uncertainty estimation method.

In this subsection, we show the equivalence between fully-connected neural networks with a proper scoring rule (log-likelihood) and the corresponding statistical model. 
We have already shown that the equivalence between dropout training and deep Gaussian process with variational approximation. In order to further formulate a fully-connected neural network with log-likelihood as a statistical model, we adds an additional generative step to deep Gaussian process that converts \eqref{eqn:gp_org} into a new statistical model,
\begin{equation}
\begin{split}
p(\mathbf{F|X}) & \sim \mathcal{N}(0, K(\mathbf{X}, \mathbf{X})), \\
p(\mathbf{Z|F}) & \sim \mathcal{N}(F, \tau^{-1}\mathbf{I}),\\
p(\mathbf{Y|Z}) & \sim g(\mathbf{Y}; \mathbf{Z}),
\end{split}
\label{eqn:gp_latent}
\end{equation}
where $g(\mathbf{Y}; \mathbf{Z})$ is a distribution that converts latent Gaussian process into predictive distribution that conforms the proper scoring rule, \ie log-likelihood. 

For regression problems, $p(\mathbf{Y|Z})$ is the Gaussian distribution,
\begin{equation}
\begin{split}
\mathbf{Z} &= [\mathbf{Z}_\mu, \mathbf{Z}_{\sigma^2}],\\
p(\mathbf{Y|Z}) &\sim \mathcal{N}(\mathbf{Z}_\mu, \mathbf{Z}_{\sigma^2}).
\end{split}
\label{eqn:gp_reg}
\end{equation}

For classification problems, $p(\mathbf{Y|Z})$ is the composition of categorical distribution with softmax function
\begin{equation}
\begin{split}
p(\mathbf{Y}_{nk}|\mathbf{Z}_{n\cdot}) &\sim \frac{\exp(Z_{nk})}{\sum_{k'}\exp(Z_{nk'})}.
\end{split}
\label{eqn:gp_clas}
\end{equation}

Therefore, the final predictive distribution estimation is changed from \eqref{eqn:gp_predict} into
\begin{equation}
p(\mathbf{y}|\mathbf{x},\mathbf{X}, \mathbf{Y}) = \int p(\mathbf{y}|\mathbf{z})\Big( \int p(\mathbf{z}|\mathbf{x}, \mathcal{W}) p(\mathcal{W}|\mathbf{X}, \mathbf{Y}) d\mathcal{W} \Big) d\mathbf{z}.
\label{eqn:gp_predict_R}
\end{equation}

In order to calculate the predictive probability \eqref{eqn:gp_predict_R}, we still have to propose the same variational distribution $q(\mathcal{W})  = \prod_{l=1}^L p(\tilde{\mathbf{W}}^{(l)})$ to approximate the posterior distribution $p(\mathcal{W}|\mathbf{X}, \mathbf{Y})$, where $\tilde{\mathbf{W}}^{(l)}$ is the random variable used in dropout operations introduced in \eqref{eqn:dropout}.

Then, in order to optimize over the variational distribution, the log evidence lower bound for the likelihood can be derived from the likelihood function,
\begin{equation}
\begin{split}
&\log p(\mathbf{Y}|\mathbf{X})\\
=&\log \int p(\mathbf{Y}|\mathbf{Z}) p(\mathbf{Z}|\mathbf{X}, \mathcal{W})p(\mathcal{W})d\mathcal{W}d\mathbf{Z}\\
=&\log \int q(\mathcal{W}) p(\mathbf{Y}|\mathbf{Z}) p(\mathbf{Z}|\mathbf{X}, \mathcal{W})\frac{p(\mathcal{W})}{q(\mathcal{W})}d\mathcal{W}d\mathbf{Z}\\
\ge& \int q(\mathcal{W}) p(\mathbf{Z}|\mathbf{X}, \mathcal{W}) \log\Bigg(  p(\mathbf{Y}|\mathbf{Z}) \frac{p(\mathcal{W})}{q(\mathcal{W})} \Bigg) d\mathcal{W}d\mathbf{Z}\\
= & \int q(\mathcal{W}) p(\mathbf{Z}|\mathbf{X}, \mathcal{W})  \log p(\mathbf{Y}|\mathbf{Z}) d\mathcal{W}d\mathbf{Z} - KL\big(q(\mathcal{W})||p(\mathcal{W})\big).
\end{split}
\label{eqn:lelb}
\end{equation}

Therefore we minimize the negative log evidence lower bound derived in \eqref{eqn:lelb} to optimize the variational parameters $\{\tilde{\mathbf{W}}^{(l)}\}$,
\begin{equation}
\mathcal{L}_{lgp} = - \sum_{n=1}^N\int p(\mathbf{z}_n|\mathbf{x}_n, \mathcal{W}) \cdot \log p(\mathbf{y}_n|\mathbf{z}_n) d\mathcal{W}d\mathbf{Z} + KL\big(q(\mathcal{W})||p(\mathcal{W})\big).
\label{eqn:nlelb}
\end{equation}

The first integral in \eqref{eqn:nlelb} can be approximated with Monte Carlo integration and the second term can be approximated according to MCDrop~\cite{Gal2016DropoutAA},
\begin{equation}
\mathcal{L}_{lgp_mc} = - \sum_{n=1}^N\log p\big(\mathbf{y}_n|\hat{\mathbf{z}}_n(\mathbf{x}_n, \hat{\mathcal{W}})\big) + \frac{p_i l^2}{2\tau N}\sum_{l=1}^L \|\mathbf{W}^{(l)}\|_2^2.
\label{eqn:nlelb_mc}
\end{equation}

Then it is trivial to verify that \eqref{eqn:nlelb_mc} is equivalent to the second equation in \eqref{equ:loss_regression} and \eqref{equ:loss_classification} for regression and classification problems respectively by substituting $p(\mathbf{y}_n|\hat{\mathbf{z}}_n(\mathbf{x}_n, \hat{\mathcal{W}}))$ with \eqref{eqn:gp_reg} or \eqref{eqn:gp_clas}.

Now, we have shown that training a fully-connected neural network with dropout and negative log-likelihood loss function is equivalent to a Gaussian or categorical distribution based on the latent deep Gaussian process.

\subsubsection{Dropout with weighted sum of negative log-likelihood and mean square error}~\label{sec:weighted}
{\color{cmtColor}
Training a neural network with a proper scoring rule, log-likelihood loss, should generate predictive uncertainty estimations that faithfully reflect the probability that the prediction will happen. However, training a neural network will log-likelihood loss solely could converge to a local optima that overestimates the true uncertainty empirically, which will be shown in our evaluation Section~\ref{sec:accuracy_uncertainty}.

The intuitive explanation for this phenomenon is straight-forward. During the early phase of training a neural network with log-likelihood loss, it is relatively hard to generate an accurate estimation of predictive mean. Then increasing the value of variance estimation can consistently decrease the negative log-likelihood loss with a high probability, since there is only a logarithm term that prevents variance from increasing as shown in~\eqref{equ:loss_regression}. Therefore, the predictive uncertainty tends to favor an estimation with large variance that overestimates the true uncertainty.
As a result, although log-likelihood loss is a proper score rule that assigns more credits to predictive uncertainties with higher quality, it usually fails to achieve a good bias-variance tradeoff during training process in practice.}

In order to achieve a well-calibrated uncertainty estimation, \ie an estimation that neither underestimates nor overestimates, we design a tunable  proper scoring rule as the training objective function of RDeepSense. It is a weighted sum of log-likelihood and negative mean square error controlled by a hyper-parameter $\alpha$,
\begin{equation}
 (1-\alpha)\cdot \log p_{\mathcal{W}}(\hat{y} = y|\mathbf{x}) - \alpha\cdot(\hat{y} - y)^2.
 \label{eqn:weighted_loss}
\end{equation}

With the definition in Section~\ref{sec:scoring}, we can easily see that \eqref{eqn:weighted_loss} is a proper scoring rule. 

According to the analysis in the previous two subsections~\ref{sec:gp} and \ref{sec:intergration}, we can see that RDeepSense, training fully-connected neural network by maximizing the weighted sum of log-likelihood and negative mean square error, is equivalent to the mixture distribution of a Gaussian or categorical distribution based on the latent deep Gaussian process and a deep Gaussian process.

{\color{cmtColor}
Since training solely with negative mean square error or log-likelihood tends to underestimate or overestimate the predictive uncertainties respectively, it is easy to fine-tune the hyper-parameter $\alpha$ with the validation dataset. When the predictive uncertainty is underestimated, we decrease the value $\alpha$, and vice versa. The detailed analysis of the effect of hyper-parameter $\alpha$ will be illustrated in Section~\ref{sec:alpha_evaluation}.}

\subsection{RDeepSense uncertainty estimation}~\label{sec:RDeepSens_run}
The previous sections prove that RDeepSense is a mathematically grounded method to estimate predictive uncertainties for fully-connected neural networks. In this section, we show that RDeepSense can efficiently estimate predictive uncertainties of fully-connected neural networks with only little computational overhead.

According to the analysis in Section~\ref{sec:intergration}, the approximated predictive distribution is
\begin{equation}
q(\mathbf{y}|\mathbf{x}) = \int p(\mathbf{y}|\mathbf{x}, \mathcal{W}) q(\mathcal{W}) d\mathcal{W} = \mathbb{E}_{q(\mathcal{W})}\big[p(\mathbf{y}|\mathbf{x}, \mathcal{W})\big],
\label{eqn:pred_dist}
\end{equation}
where $\mathcal{W} =\{\tilde{\mathbf{W}}^{(l)}\}$ is the random variables generated by dropout operations at each layer.
\begin{equation}
\begin{split}
\mathbf{z}_{[i]}^{(l)} &\sim  \text{Bernoulli}(\mathbf{p}_{[i]}^{(l)}), \\
\tilde{\mathbf{W}}^{(l)} &=  \text{diag}\big( \mathbf{z}^{(l)}\big) \mathbf{W}^{(l)}.
\end{split}
\label{eqn:approx_W}
\end{equation}

Usually Monte Carlo estimation is used to approximate the predictive distribution $q(\mathbf{y}|\mathbf{x})$ through sampling random variables $\mathcal{W}$,
\begin{equation}
q(\mathbf{y}|\mathbf{x})  = \frac{1}{M}\sum_{m=1}^M p(\mathbf{y}|\mathbf{x}, \mathcal{W}_m).
\label{eqn:pred_dist_MC}
\end{equation}

{\color{cmtColor}
For classification, \eqref{eqn:pred_dist_MC} is the average of categorical distribution. For regression, \eqref{eqn:pred_dist_MC} is an average of Gaussian distributions. If we assume that $M$ Gaussian distributions are independent, the resulted average distribution can be approximated by a single Gaussian distribution according to the central limit theorem,}
\begin{equation}
\begin{split}
\frac{1}{M} \sum_{m=1}^M p(\mathbf{y}|\mathbf{x}, \mathcal{W}_m) &= \sum_{m=1}^M\mathcal{N}(\mu_m(\mathbf{x}), \sigma_m^2(\mathbf{x}))\\
 & = \mathcal{N}(\hat{\mu}(\mathbf{x}), \hat{\sigma}^2(\mathbf{x})),\\
 \hat{\mu}(\mathbf{x}) &= \frac{1}{M}\sum_{m=1}^M \mu_m(\mathbf{x}),\\
 \hat{\sigma}^2(\mathbf{x}) &= \frac{1}{M}\sum_{m=1}^M \big(\sigma_m^2(\mathbf{x}) + \mu_m^2(\mathbf{x})\big) -  \hat{\mu}^2(\mathbf{x}).
 \end{split}
\label{eqn:pred_dist_reg}
\end{equation}

The drawback of Monte Carlo estimation for embedded devices is its high energy and time consumptions. We have to run the whole neural network for $M$ times to generate $M$ samples, which is not suitable for embedded devices with limited resources.

Fortunately, there is a simple yet effective recipe proposed by the dropout operation that can effectively approximate the expected output value instead of using Monte Carlo estimation~\cite{srivastava2014dropout}. During test time, the dropout operation is changed from~\eqref{eqn:dropout} into
\begin{equation}
\begin{split}
\tilde{\mathbf{W}}^{(l)} &=  \text{diag}\big( \mathbf{p}^{(l)}\big) \mathbf{W}^{(l)}, \\
\mathbf{y}^{(l)} &= \mathbf{x}^{(l)} \tilde{\mathbf{W}}^{(l)}  + \mathbf{b}^{(l)},\\
\mathbf{x}^{(l+1)} &= f^{(l)}\big(\mathbf{y}^{(l)}\big).
\end{split}
\label{eqn:dropout_test}
\end{equation}

{\color{cmtColor}
Although the approximation~\eqref{eqn:dropout_test} is not theoretically equivalent to the Monte Carlo estimation~\eqref{eqn:pred_dist_reg} by assuming the zero variance of mean estimation, $\sum_{m=1}^M\mu_m^2(\mathbf{x}) - (\sum_{m=1}^M\mu_m(\mathbf{x}))^2 = 0$, the proposed approximation \eqref{eqn:dropout_test} turns to be an effective and efficient approximation during the evaluation in Section~\ref{sec:evaluation}. In the evaluation section, we will empirically compare the biased approximation~\eqref{eqn:dropout_test} with the unbiased Monte Carlo estimation~\eqref{eqn:pred_dist_reg}.}

Therefore, with the approximation~\eqref{eqn:dropout_test}, we can directly estimate the expected predictive mean and variance of a Gaussian distribution for regression problems and expected categorical probabilities for classification problems by just running the neural network for a single time. This makes RDeepSense a suitable candidate for deep neural networks with uncertainty estimations used in mobile and ubiquitous computing applications.

\section{Evaluation}\label{sec:evaluation}
In this section, we evaluate RDeepSense on four mobile and ubiquitous computing tasks. We first introduce the experimental setup for each task, including hardware, datasets, and baseline algorithms. 
We then evaluate the accuracy and the quality of uncertainty estimation. 
Next we evaluate the inference time and energy consumption of all algorithms on the testing hardware.
At last we evaluate and analyze the effect of hyper-parameter $\alpha$ in the training objective function~\eqref{eqn:weighted_loss} on the model performance such as accuracy and quality of uncertainty estimation.

\subsection{Testing hardware}
Our testing hardware is based on Intel Edison computing platform~\cite{Edison}. The Intel Edison computing platform is powered by the Intel Atom SoC dual-core CPU at 500 MHz and is equipped with 1GB memory and 4GB flash storage. For fairness, all neural network models are run solely on CPU during evaluation for inference time and energy consumption.

\subsection{Evaluation tasks}~\label{sec:eval_task}
We conduct four experiments related to human health and wellbeing, smart city transportation, environment monitoring, and human activity recognition with RDeepSense and other two state-of-the-art deep learning uncertainty measuring methods as well as a statistical model. The experimental settings of the tasks and datasets are introduced in this subsection.

The detailed statistical information of four datasets is illustrated in Table~\ref{tab:dataset}
\begin{table}[!htb]
%\vspace{-0.3cm}
\small
\begin{center}
\caption {Statistical Information of four datasets used in evaluations}
\vspace{-0.1cm}
\label{tab:dataset}
\begin{tabular}{ |c | c | c | c | c | c | c | } 
 \hline
 Dataset & Training Size & Validating Size  &  Testing Size & Mean of output & Std of output & Range of output  \\ 
  \hline
   \hline
  BPEst & 1,281,098 & 26,689 & 26,689 & 88.74 & 25.01 & $[50.0, 199.93]$ \\ 
  \hline
  NYCommute &10,287,766 & 214,328 & 214,328 & 15.08 & 52.79 & $[0.0, 1439.5]$ \\ 
 \hline
 GasSen & 2,839,933 & 59,166 & 59,166 & 94.56 & 145.16 & $[0.0, 533.33]$ \\ 
 \hline
 HHAR & 28,314 & 1,686 & 1,686 & N/A & N/A & $\{0,1,2,3,4,5\}$ \\ 
 \hline
\end{tabular}
\end{center}
%\vspace{-0.6cm}
\end{table} 

\begin{itemize}
\item\textit{BPEst: Cuff­less blood pressure monitoring through photoplethysmogram.} 
The first task is to monitor cuff­less blood pressure through photoplethysmogram from fingertip. The dataset is originally collected by patient monitors at various hospitals between 2001 and 2008. Waveform signals were sampled at the frequency of 125 Hz with at least 8 bit accuracy~\cite{goldberger2000physiobank}. The photoplethysmogram from fingertip (PPG) and arterial blood pressure (ABP) signal (mmHg) is extracted by Mohamad et al. for the non-invasive cuff­less blood pressure monitoring task~\cite{kachuee2015cuff}.\footnote{\url{https://archive.ics.uci.edu/ml/datasets/Cuff-Less+Blood+Pressure+Estimation}} 
The target of BPEst task is to infer the waveform of  ABP based on the waveform of PPG collected from fingertips. This is a more challenging task compared with estimating the upper and lower bound of the blood pressure, which requires a more precise estimation of predictive uncertainty. During the experiment, a learning model is trained to estimate a 2-second ABP waveform (250 samples) based on the corresponding 2-second PPG waveform.

\item\textit{NYCommute: Commute time estimation of New York City.}
Smart transportation is an increasingly important task within the topic of smart city.
The second task is to estimate commute time in New York City through the pick-up time and location as well as the drop-off location. We use the yellow and green taxi trip records within January 2017 as the training, validation, and testing dataset.\footnote{\url{http://www.nyc.gov/html/tlc/html/about/trip_record_data.shtml}} The input of the learning model is a vector with 5 elements, containing the standardized longitude and latitude of pick-up and drop-off location as well as the pick-up time within a day. The output of the learning model is the expected commute time and its corresponding uncertainty estimation.

\item\textit{GasSen: Estimate dynamic gas mixtures from chemical sensors.}
The third task is related to the environment monitoring. The task is to estimate real concentration of Ethylene and CO gas mixture from an array of low-end chemical sensors. Fonollosa et al. constructed the dataset by the continuous acquisition the signals of a sensor array with 16  chemical sensors for a duration of about 12 hours without interruption with the sampling frequency of 100 Hz~\cite{fonollosa2015reservoir}.~\footnote{\url{https://archive.ics.uci.edu/ml/datasets/Gas+sensor+array+under+dynamic+gas+mixtures}} Gas concentrations range from $0-­600$ parts-per-million (ppm). The learning model is trained and tested to predict the concentration Ethylene and CO gas mixtures through the vector of 16 sensor inputs.

\item\textit{HHAR: Heterogeneous human activity recognition.}
The previous three tasks are all regression tasks, but this one is a classification task. 
Heterogeneous means that we are testing on a new user who has not appeared in the training set. This dataset contains readings from two motion sensors (accelerometer and gyroscope). Readings are recorded when users execute activities scripted in no specific order, while carrying smartwatches and smartphones. The dataset contains 9 users, 6 activities (biking, sitting, standing, walking, climbStairup, and climbStair-down), and 6 types of mobile devices~\cite{stisen2015smart}~\footnote{\url{https://archive.ics.uci.edu/ml/datasets/Heterogeneity+Activity+Recognition}}. 
{\color{cmtColor}
We segment raw measurements into 5-second samples and take Fourier transform on these samples as the input data.
For future extension to RNNs as discussed in Section~\ref{sec:discussion}, each sample is further divided
into time intervals of length $\tau=0.25$s. Then we calculate the frequency response of sensors for
each time interval.}
The output of HHAR is one of the 6 activities.
\end{itemize}

\subsection{Baseline algorithms}~\label{sec:baseline_algorithm}
We compare RDeepSense with other two state-of-the-algorithm deep learning uncertainty estimation algorithms, {\color{cmtColor}RDeepSense with Monte Carlo estimation}, and Gaussian process. The algorithms with deep neural network, including RDeepSense, use the same neural network architecture. It is a 4-layer fully-connected neural network with $500$ hidden dimension. 

\begin{itemize}
\item\textit{MCDrop:} {\color{cmtColor}This algorithm is based on the Monte Carlo dropout as described in Section~\ref{sec:gp}~\cite{Gal2016DropoutAA}.} Compared with RDeepSense, the main difference is that MCDrop is not optimized by a proper scoring rule. MCDrop requires running the neural network for multiple times to generate samples during uncertainty estimation. Therefore we use MCDrop-k to represent MCDrop with $k$ samples. {\color{cmtColor}Multiple samples, \ie $k>1$, are required to generate a predictive uncertainty estimation.} During the evaluation, we let $k$ to be $3$, $5$, $10$, and $20$ to evaluate the tradeoff between the quality of uncertainty estimation and the resource consumption for MCDrop. 

\item\textit{SSP:} {\color{cmtColor}This algorithm trains the neural network with proper scoring methods and uses the ensemble method~\cite{lakshminarayanan2016simple}.} Compared with RDeepSense, the main difference is that SSP uses the ensemble method instead of the dropout operation in each layer. SSP requires training multiple neural networks for ensemble. Therefore we use SSP-k to represent SSP by ensemble $k$ individual neural networks. During the evaluation, we let $k$ to be $1$, $3$, $5$, and $10$ to evaluate the tradeoff between the quality of uncertainty estimation and the resource consumption for SSP. 

\item{\color{cmtColor}\textit{RDeepSense-MC:} This algorithm is basically the proposed RDeepSense algorithm. The difference is that, during the inference, RDeepSense-MC uses Monte Carlo estimation~\eqref{eqn:pred_dist_reg} instead of the efficient approximation~\eqref{eqn:dropout_test} for uncertainty estimation. Therefore we use RDeepSense-MCk to present RDeepSense-MC with $k$ samples. During the evaluation, we let $k$ to be $3$, $5$, $10$, and $20$ to evaluate the effectiveness and efficiency of RDeepSense inference approximation~\eqref{eqn:dropout_test} compared with the Monte Carlo estimation~\eqref{eqn:pred_dist_reg}.}

\item\textit{GP:} Gaussian process (GP) is the baseline algorithm used during the evaluation of accuracy and the quality of uncertainty estimations, but not for the evaluations of running time and energy consumption on Edison. The main reason is that the computation cost during model inference for GP is $O(N^3)$, where $N$ is the number of data instances. This cost can be prohibitive even for moderately sized datasets on embedded devices, such as Intel Edison. In additional, GP requires $O(N^2)$ memory consumption during training. Therefore we train the GP with only a proportion of dataset on a server with 128GB memory. Notice that GP is the baseline used to illustrate the quality of uncertainty estimations generated by a statistical model, so the size of training dataset is not the main concern.
\end{itemize}

\subsection{Accuracy of prediction and quality of uncertainty estimations}~\label{sec:accuracy_uncertainty}
In this section, we discuss the accuracy and the uncertainty estimation quality of RDeepSense compared with the other baseline algorithms. {\color{cmtColor}RDeepSense is tuned with the validating dataset, and all algorithms in all experiments are tested on the testing dataset.}

For three regression problems, two types of evaluation results will be illustrated and discussed. The first type of evaluation is based on some basic measurements including mean absolute error  and negative log-likelihood. The second type of evaluation is based on the calibration curves, also known as reliability diagrams. We compute the $z\%$ confidence interval for each testing data based on predictive mean and variance of each algorithm. Then we measure the fraction of the testing data that falls into this confidence interval. For a well-calibrated uncertainty estimation, the fraction of testing data that falls into the confidence interval should be similar to $z\%$. We compute the calibration curves with $z = [10\%, 20\%, 30\%, 40\%, 50\%, 60\%, 70\%, 80\%, 85\%, 95\%, 99\%, 99.5\%, 99.9\%]$ for all three regression problems. 

For the classification problem, the calibration curve is not available. Therefore, we evaluate HHAR  based on accuracy, {\color{cmtColor}f1 score}, negative log-likelihood, and a new measurement called the mean entropy of false predictions. If the entropies of false predictions are higher, the learning algorithms show more uncertainties about the false predictions, which represents a better quality of uncertainty estimations.

\begin{table}[!htb]
%\hspace{-1.3cm}
\small
\begin{center}
\caption {{\color{cmtColor}Mean Absolute Error (MAE) and Negative Log-Likelihood (NLL) for the BPEst task. Except for RDeepSense-MC20, RDeepSense is the best-performing algorithm for NLL and is the second best-performing algorithm for MAE.}}
%\vspace{-0.1cm}
\label{tab:exp_BPEst}
\begin{tabular}{ |c | c | c | c | c | c| } 
 \hline
 & RDeepSense  & {\color{cmtColor}RDeepSense-MC3} & {\color{cmtColor}RDeepSense-MC5}  &  {\color{cmtColor}RDeepSense-MC10} & {\color{cmtColor}RDeepSense-MC20}  \\
 \hline
  MAE & $\mathbf{14.18}$ & $14.93$ & $14.64$ & $14.44$ & $14.32$  \\ 
  \hline
  NLL & $\mathbf{3.46}$ & $3.49$ & $3.47$ & $3.46$  &  $\mathbf{3.45}$  \\ 
 \hline
  \hline
 & SSP-1  & SSP-3 & SSP-5  &  SSP-10 & GP \\
  \hline
  MAE &  $15.76$ & $14.68$ &$14.67$ & $14.78$ & $19.15$  \\ 
  \hline
  NLL &  $4.4$ & $3.69$ & $3.48$ & $3.49$ & $3.59$  \\ 
 \hline
 \hline
 & MCDrop-3  & MCDrop-5 & MCDrop-10  &  MCDrop-20 &  \\
  \hline
   MAE & $14.80$ & $14.41$& $\mathbf{14.09}$ & $\mathbf{14.09}$ &  \\ 
  \hline
  NLL & $38.1$ & $5.28$ & $4.00$ & $4.00$ &  \\ 
 \hline
\end{tabular}
\end{center}
%\vspace{-0.6cm}
\end{table} 

\begin{figure}[!htb]
\vspace{-0.3cm}
\begin{subfigure}{.4\linewidth}
  \centering
  \includegraphics[width=1.\linewidth]{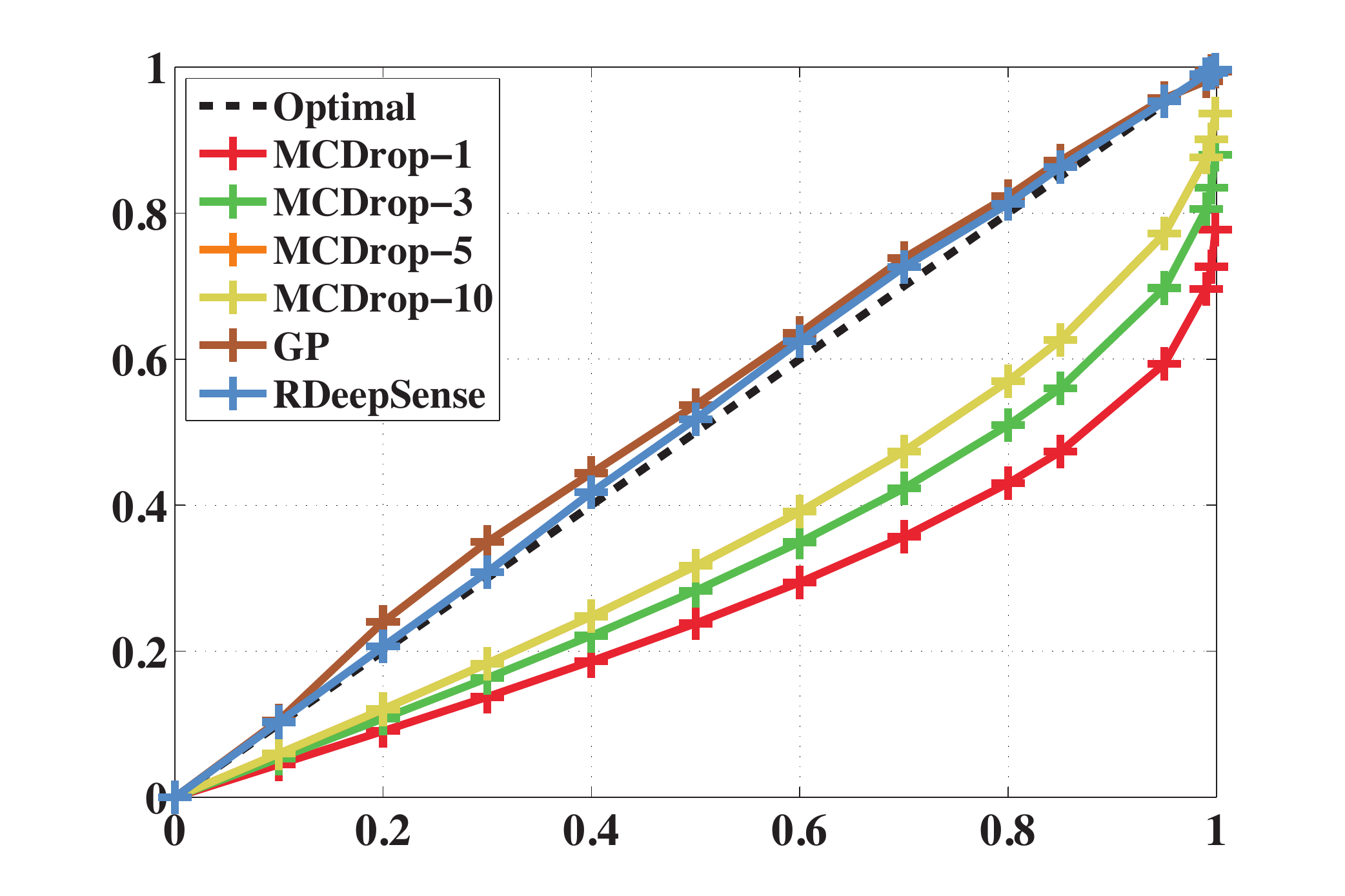}
  \caption{The calibration curves of RDeepSense, GP, and MCDrop-k.}
  \label{fig:BPEst_conf_MCDrop}
\end{subfigure}
\begin{subfigure}{.4\linewidth}
  \centering
  \includegraphics[width=1.\linewidth]{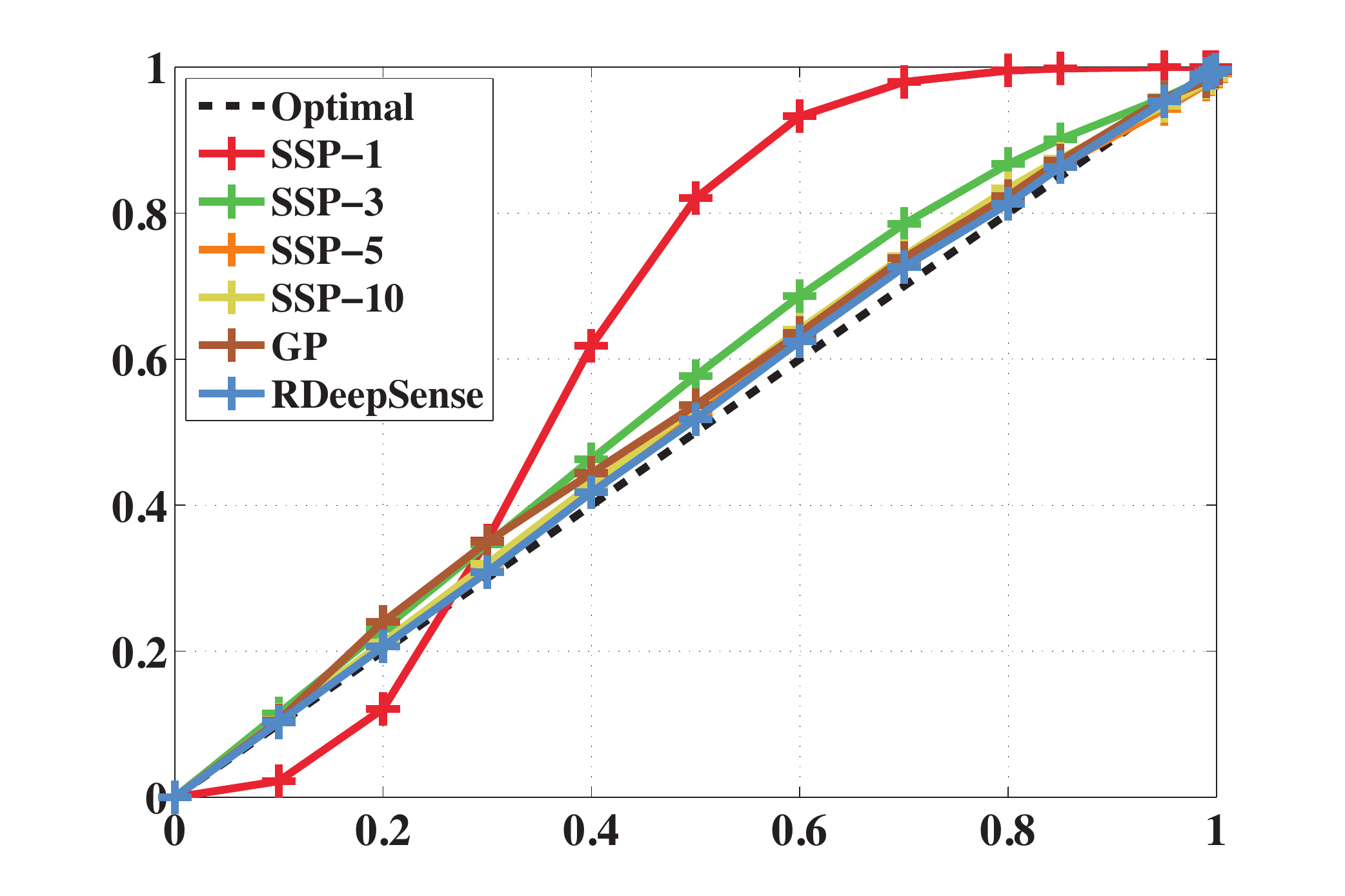}
  \caption{The calibration curves of RDeepSense, GP, and SSP-k.}
  \label{fig:BPEst_conf_SSP}
\end{subfigure}
\begin{subfigure}{.4\linewidth}
  \centering
  \includegraphics[width=1.\linewidth]{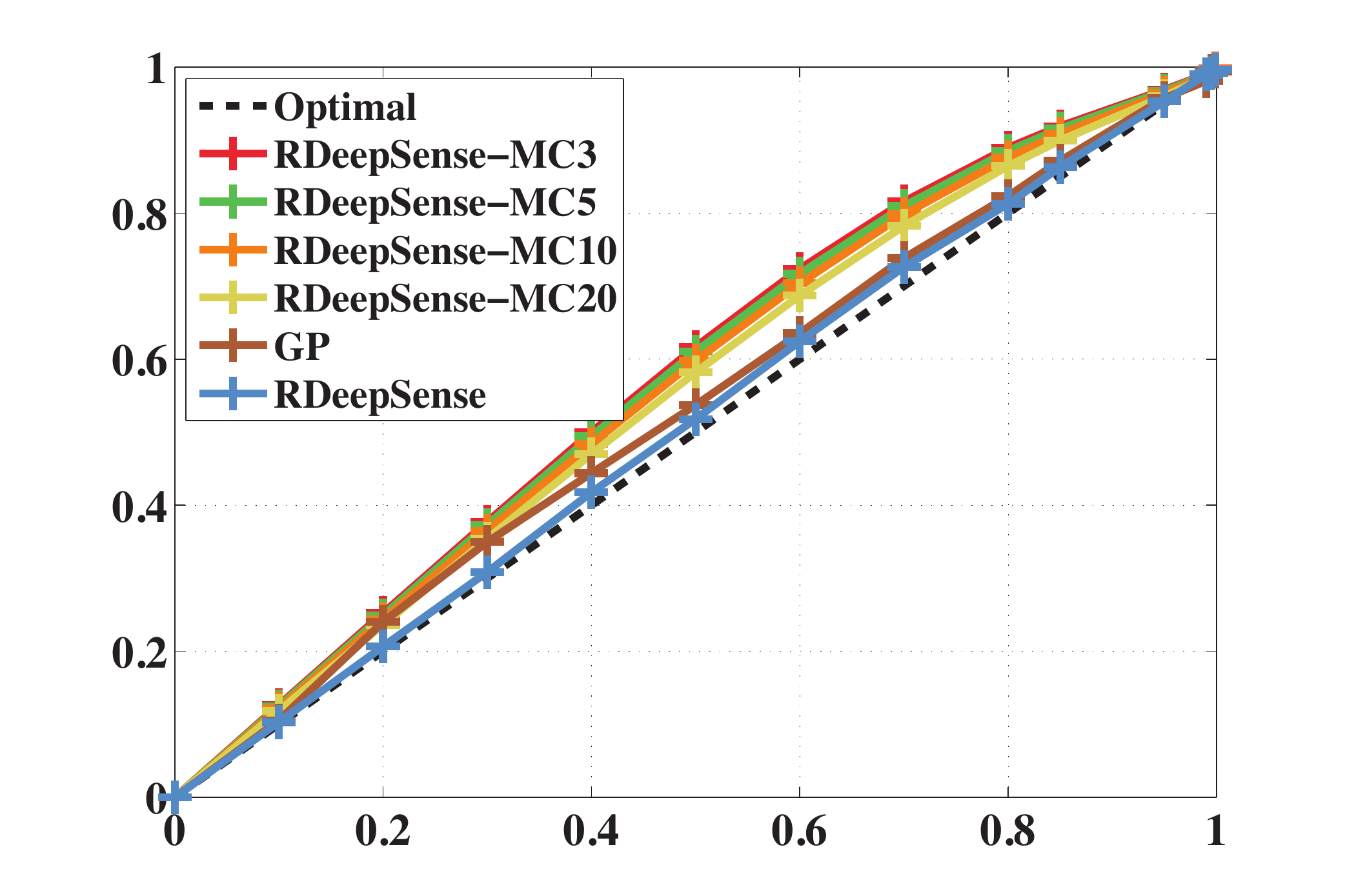}
  \caption{{\color{cmtColor}The calibration curves of RDeepSense, GP, and RDeepSense-MCk.}}
  \label{fig:BPEst_conf_RDeepSense_MC}
\end{subfigure}
\caption{{\color{cmtColor}The calibration curves of BPEst for RDeepSense, GP, MCDrop-k, SSP-k, and RDeepSense-MCk. MCDrop-k underestimates the predictive distribution. SSP-k overestimates the predictive distribution. RDeepSense is the closest curve to the optimal predictive distribution.}}
\label{fig:BPEst_conf}
%\vspace{-0.15cm}
\end{figure}

\subsubsection{BPEst} We first compare RDeepSense with four baseline algorithms based on mean absolute error (MAE) and negative log-likelihood (NLL), which is illustrated in Table~\ref{tab:exp_BPEst}, where we highlight the results of RDeepSense and the best-performing one.

From Table~\ref{tab:exp_BPEst}, we can see that, except for RDeepSense-MC20, RDeepSense is the best-performing and the second best-performing algorithm for NLL and MAE respectively, which means that RDeepSense can provide accurate estimation with high-quality predictive uncertainty. 
{\color{cmtColor}RDeepSense-MC20 only slightly beats RDeepSense on NLL, however RDeepSense-MC20 consumes around $\times 20$ time and energy compared with RDeepSense.}
The performance of MCDrop-k increases when k increases. Larger k means that MCDrop algorithm generates more samples during model inference, which can provide higher-quality estimations but more resource consumptions. MCDrop-3 provides a relatively bad result for NLL, which means MCDrop does require a number of samples for uncertainty estimation with reasonable quality. The ensemble method used in SSP increases the prediction performance, but it is not consistent. SSP-10 observes the performance degradation compared with SSP-5. GP obtains a relatively large MAE. This is because GP cannot be scaled to train on the whole dataset.

The calibration curves of BPEst task is illustrated in Figure~\ref{fig:BPEst_conf}. These three figures show the quality of predictive uncertainty estimations. RDeepSense generates predictive uncertainties with the highest quality. RDeepSense even slightly out-performs the traditional statistical model, GP.  As we mentioned in Section~\ref{sec:analysis}, MCDrop-k tends to underestimate the predictive uncertainty, while SSP-k tends to overestimate the predictive uncertainty. 
{\color{cmtColor}RDeepSense even generates predictive uncertainty with better calibration compared with RDeepSense-MCk, which indicate the effectiveness of approximation during inference.}
All MCDrop-k, SSP-k, and RDeepSense-MCk improve the quality of uncertainty estimations by increasing the value of $k$.

\subsubsection{NYCommute} Then we compare RDeepSene with baseline algorithms for NYCommute task. The comparison based on Mean Absolute Error (MAE) and Negative Log-Likelihood (NLL) is shown in Table~\ref{tab:exp_NYCommute}.

\begin{table}[!htb]
%\hspace{-1.3cm}
\small
\begin{center}
\caption {{\color{cmtColor}Mean Absolute Error (MAE) and Negative Log-Likelihood (NLL) for the NYCommute task.}}
%\vspace{-0.1cm}
\label{tab:exp_NYCommute}
\begin{tabular}{ |c | c | c | c | c | c| } 
 \hline
 & RDeepSense  & {\color{cmtColor}RDeepSense-MC3} & {\color{cmtColor}RDeepSense-MC5}  &  {\color{cmtColor}RDeepSense-MC10} & {\color{cmtColor}RDeepSense-MC20}  \\
 \hline
  MAE & $\mathbf{5.64}$ & $6.10$ & $6.04$ & $5.99$ & $5.96$  \\ 
  \hline
  NLL & $\mathbf{7.7}$ & $7.85$ & $7.81$ & $7.73$  &  $\mathbf{7.7}$  \\ 
 \hline
  \hline
 & SSP-1  & SSP-3 & SSP-5  &  SSP-10 & GP \\
  \hline
  MAE &  $8.15$ & $7.90$ &$7.51$ & $7.03$ & $11.84$   \\ 
  \hline
  NLL &  $4.86$ & $\mathbf{4.67}$ & $4.84$ & $4.81$ & $7.46$  \\ 
 \hline
 \hline
 & MCDrop-3  & MCDrop-5 & MCDrop-10  &  MCDrop-20 &  \\
  \hline
   MAE & $5.69$ & $5.64$& $\mathbf{5.61}$ & $\mathbf{5.61}$ &  \\ 
  \hline
  NLL & $19995.6$ & $1335.73$ & $640.35$ & $640.35$ &  \\ 
 \hline
\end{tabular}
\end{center}
%\vspace{-0.6cm}
\end{table} 

\begin{figure}[!htb]
%\vspace{-0.3cm}
\begin{subfigure}{.4\linewidth}
  \centering
  \includegraphics[width=1.\linewidth]{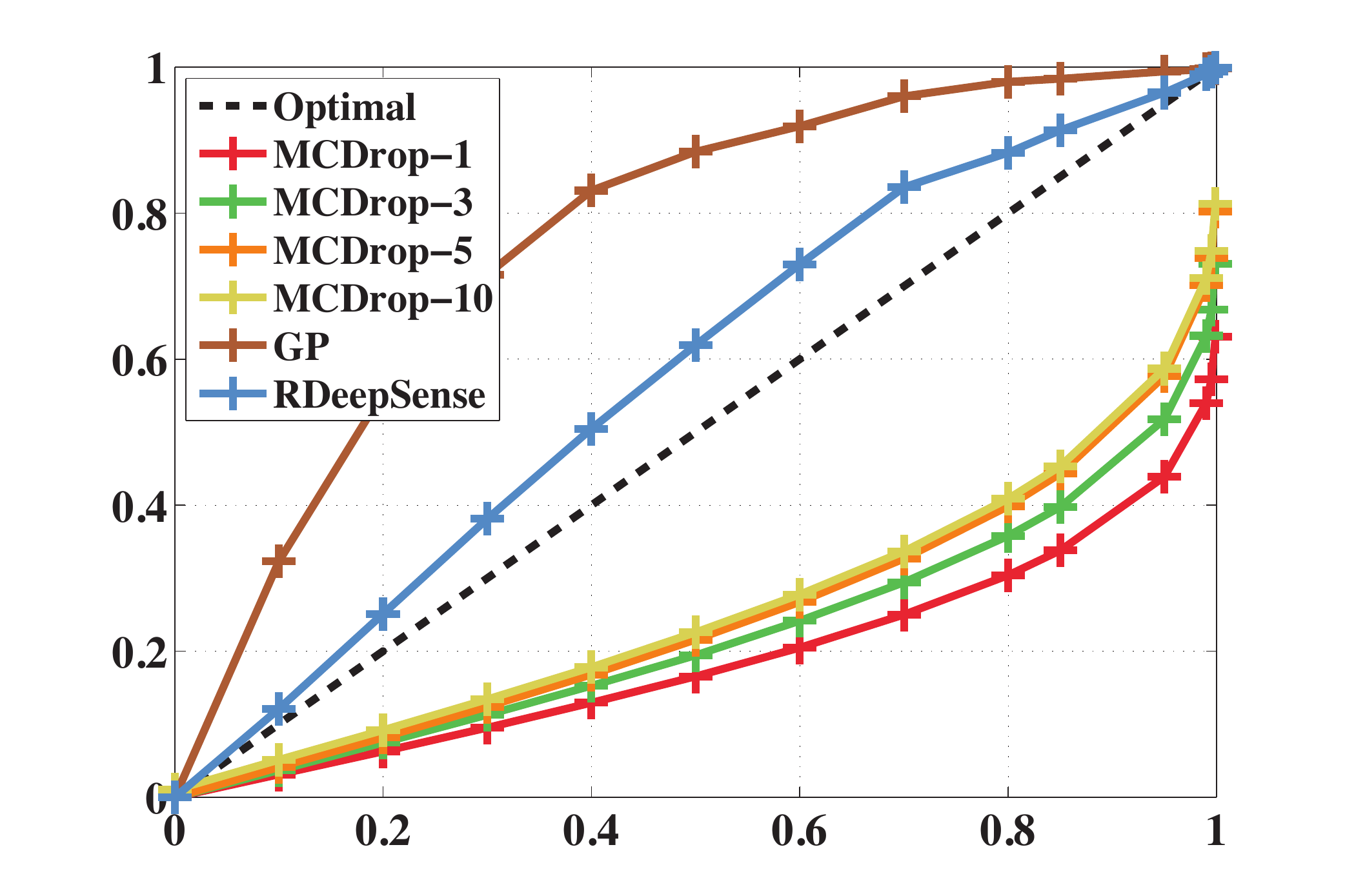}
  \caption{The calibration curves of RDeepSense, GP, and MCDrop-k.}
  \label{fig:NYCommute_conf_MCDrop}
\end{subfigure}
\begin{subfigure}{.4\linewidth}
  \centering
  \includegraphics[width=1.\linewidth]{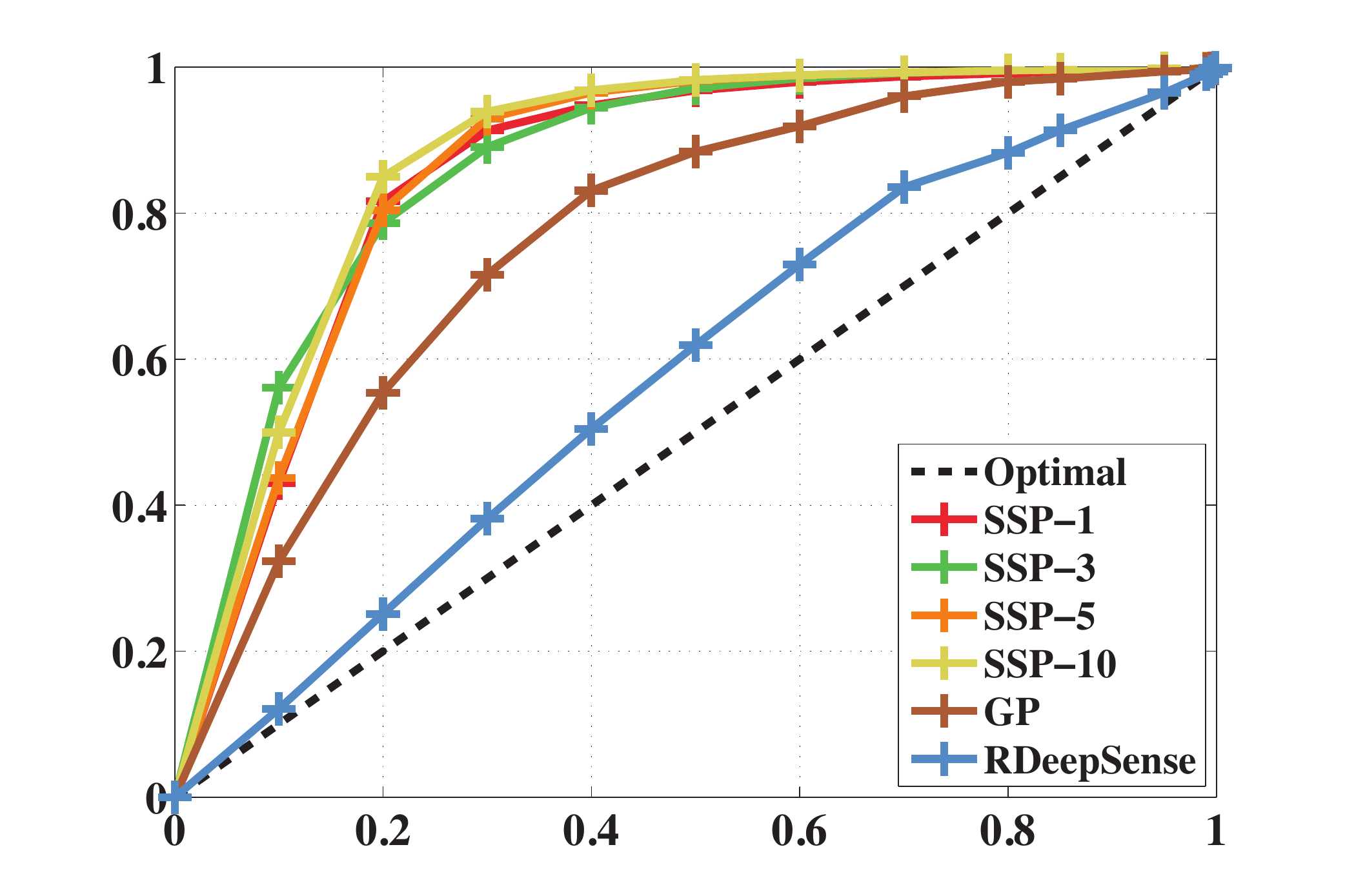}
  \caption{The calibration curves of RDeepSense, GP, and SSP-k.}
  \label{fig:NYCommute_conf_SSP}
\end{subfigure}
\begin{subfigure}{.4\linewidth}
  \centering
  \includegraphics[width=1.\linewidth]{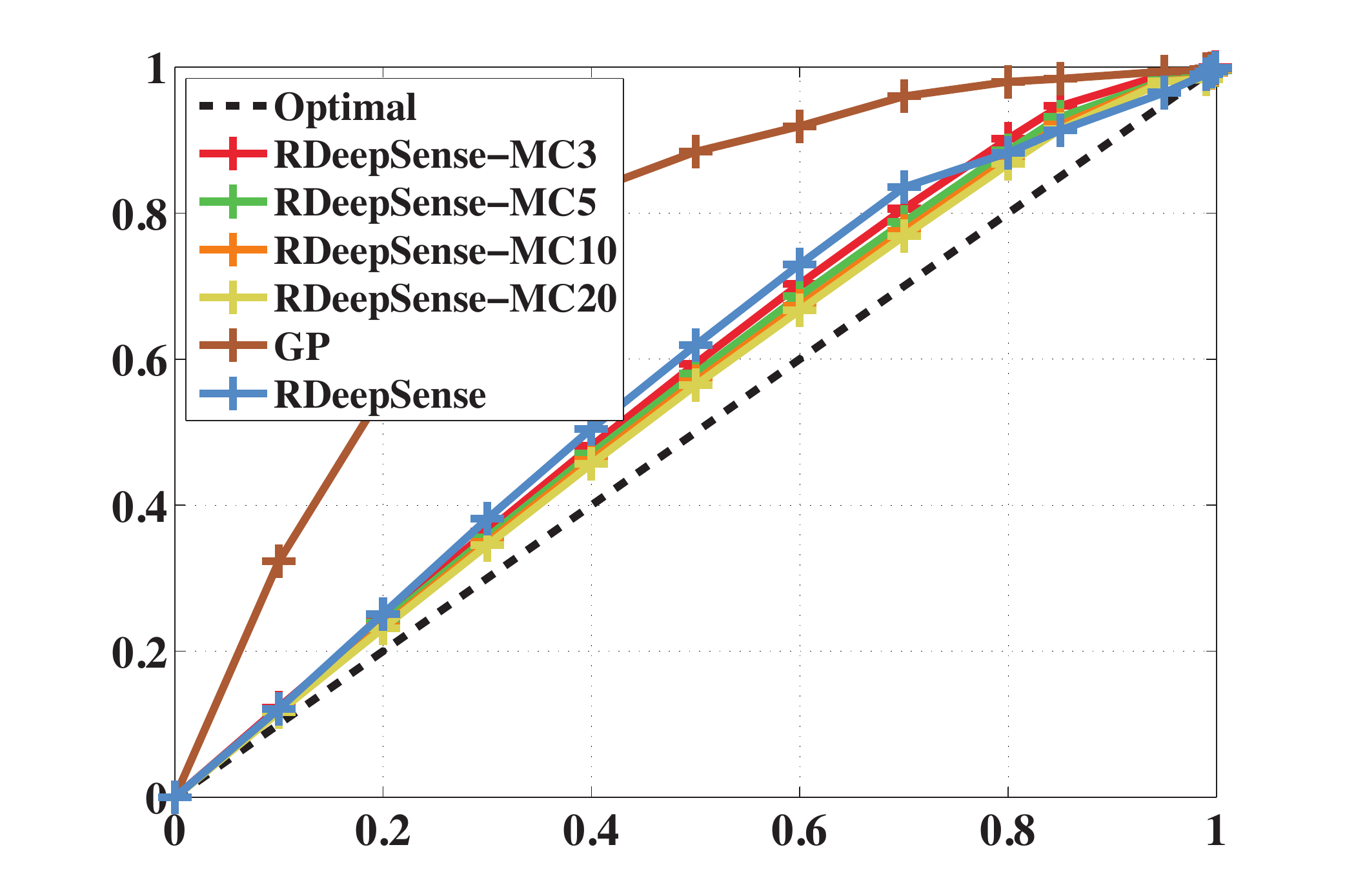}
  \caption{{\color{cmtColor}The calibration curves of RDeepSense, GP, and RDeepSense-MCk.}}
  \label{fig:BPEst_conf_RDeepSense_MC}
\end{subfigure}
\caption{{\color{cmtColor}The calibration curves of NYCommute for RDeepSense, GP, MCDrop-k, SSP-k, and RDeepSense-MCk. MCDrop-k highly underestimates the predictive distribution. SSP-k highly overestimates the predictive distribution. RDeepSense makes a tradeoff between these two and is the closest curve to the optimal predictive distribution.}}
\label{fig:NYCommute_conf}
%\vspace{-0.15cm}
\end{figure}

In this task, RDeepSense tends to find a balance between MAE and NLL measurements. MCDrop-k shows low MAE and high NLL, while SSP-k shows high MAE and low NLL. MCDrop-k tries to minimize the mean square error, while SSP-k tries to minimize the negative log-likelihood. Therefore, MCDrop-k focuses more on the mean of predictive distribution, and SSP-k focuses more on the overall likelihood. RDeepSense combines two objective functions, mean square error and negative log-likelihood, which tries to find a balance point between these two. Still, due to the scalability problem, GP obtains a relatively larger MAE.  {\color{cmtColor}Compared with RDeepSense-MCk, RDeepSense achieve a good performance on both MAE and NLL. Only RDeepSense-MC20 shows the same performance on the NLL measurement.}

The calibration curves of NYCommute task is illustrated in Figure~\ref{fig:NYCommute_conf}. Both MCDrop-k and SSP-k fail to generate high-quality uncertainty estimations by either underestimating or overestimating the predictive uncertainties. However, RDeepSense can still provide uncertainty estimations with good quality, which outperforms GP with a significant margin. {\color{cmtColor}Compared with RDeepSense-MCk, RDeepSense shows similar performance on generating well-calibrated predictive uncertainties, which shows that the approximation~\eqref{eqn:dropout_test} works well in practice.}

\subsubsection{GasSen}~\label{sec:GasSen_exp}
Next we compare RDeepSense with other baseline algorithms for the GasSen task. Table~\ref{tab:exp_GasSen} illustrates the performance of all these algorithms based on Mean Absolute Error (MAE) and Negative Log-Likelihood (NLL). Except for RDeepSense-MC20, RDeepSense is the best-performing algorithm according to these two metrics.  Similarly, MCDrop-k shows low MAE and NLL, while SSP-k shows high MAE and NLL. This is due to the objective of these two types of algorithms. MCDrop-k minimizes the mean square error, while SSP-k minimizes the negative log-likelihood. Therefore, MCDrop-k focuses more on the mean of predictive distribution, and SSP-k focuses more on the overall likelihood. RDeepSense combines two objective function. Therefore, RDeepSense is able to achieve the best performance in both cases. {\color{cmtColor}The usage of dropout that prevents feature co-adapting is the main reason why RDeepSense achieves better NLL compared with SPP-k. The RDeepSense still achieves good performance compared with its Motel Carlo version. Only RDeepSense-MC20 slightly outperforms RDeepSense under the NLL measurement, which shows the effectiveness of the approximation used in RDeepSense.}

\begin{table}[!htb]
%\hspace{-1.3cm}
\small
\begin{center}
\caption {{\color{cmtColor}Mean Absolute Error (MAE) and Negative Log-Likelihood (NLL) for the GasSen task. Except for RDeepSense-MC20, RDeepSense is the best-performing algorithm for both MAE and NLL.}}
%\vspace{-0.1cm}
\label{tab:exp_GasSen}
\begin{tabular}{ |c | c | c | c | c | c| } 
 \hline
 & RDeepSense  & {\color{cmtColor}RDeepSense-MC3} & {\color{cmtColor}RDeepSense-MC5}  &  {\color{cmtColor}RDeepSense-MC10} & {\color{cmtColor}RDeepSense-MC20}  \\
 \hline
  MAE & $\mathbf{15.25}$ & $17.21$ & $16.44$ & $16.34$ & $15.61$  \\ 
  \hline
  NLL & $\mathbf{3.77}$ & $4.23$ & $4.18$ & $3.88$  &  $\mathbf{3.73}$  \\ 
 \hline
  \hline
 & SSP-1  & SSP-3 & SSP-5  &  SSP-10 & GP \\
  \hline
  MAE &  $24.40$ & $22.53$ &$20.75$ & $20.68$ & $35.74$    \\ 
  \hline
  NLL &  $4.76$ & $4.34$ & $3.92$ & $3.81$ & $7.76$  \\ 
 \hline
 \hline
 & MCDrop-3  & MCDrop-5 & MCDrop-10  &  MCDrop-20 &  \\
  \hline
   MAE & $21.23$ & $20.45$& $19.79$ & $19.79$ &  \\ 
  \hline
  NLL & $2201.95$ & $463.94$ & $170.45$ & $170.45$ &  \\ 
 \hline
\end{tabular}
\end{center}
%\vspace{-0.6cm}
\end{table} 

\begin{figure}[!htb]
%\vspace{-0.3cm}
\begin{subfigure}{.4\linewidth}
  \centering
  \includegraphics[width=1.\linewidth]{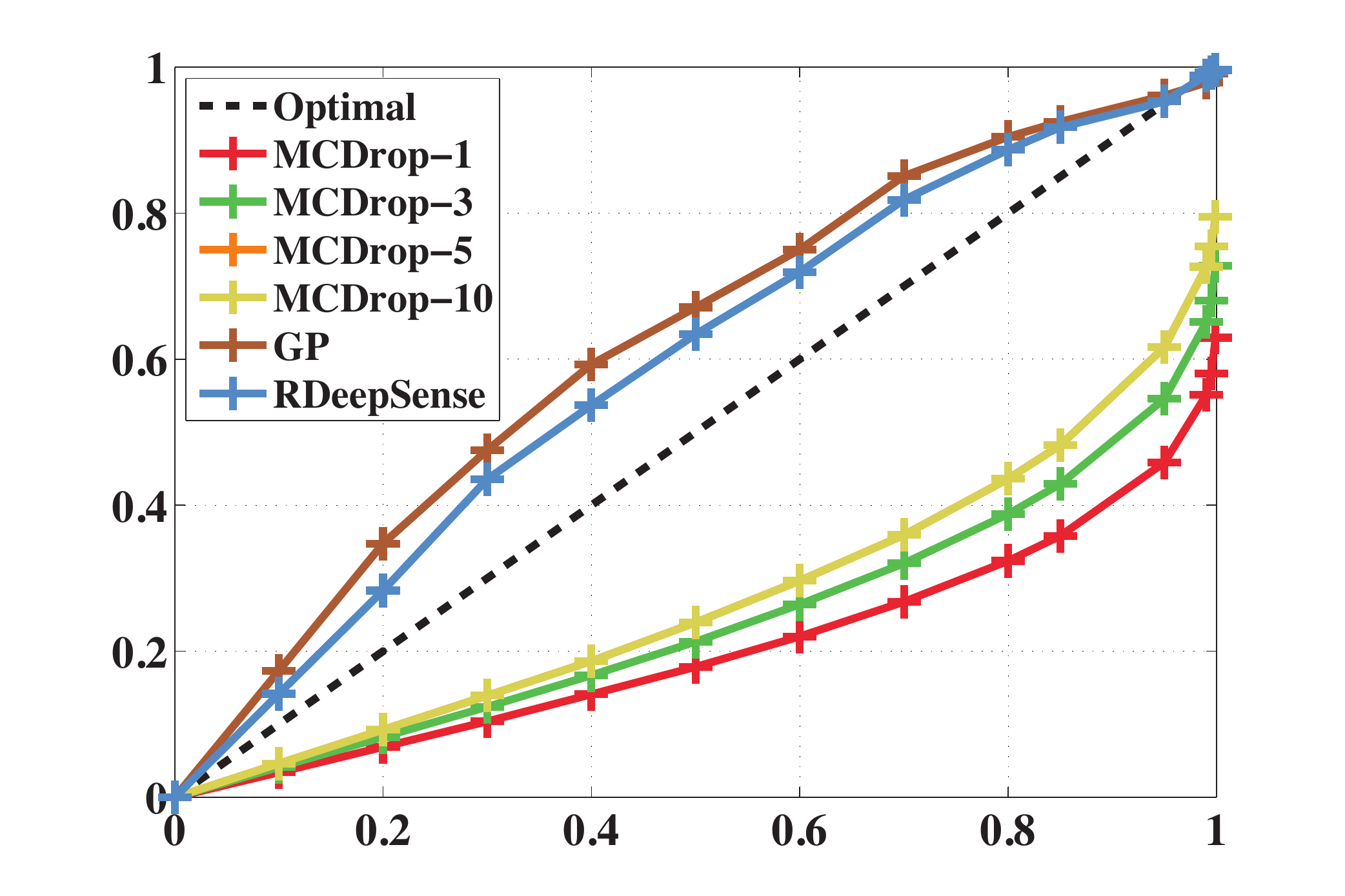}
  \caption{The calibration curves of RDeepSense, GP, and MCDrop-k.}
  \label{fig:GasSen_conf_MCDrop}
\end{subfigure}
\begin{subfigure}{.4\linewidth}
  \centering
  \includegraphics[width=1.\linewidth]{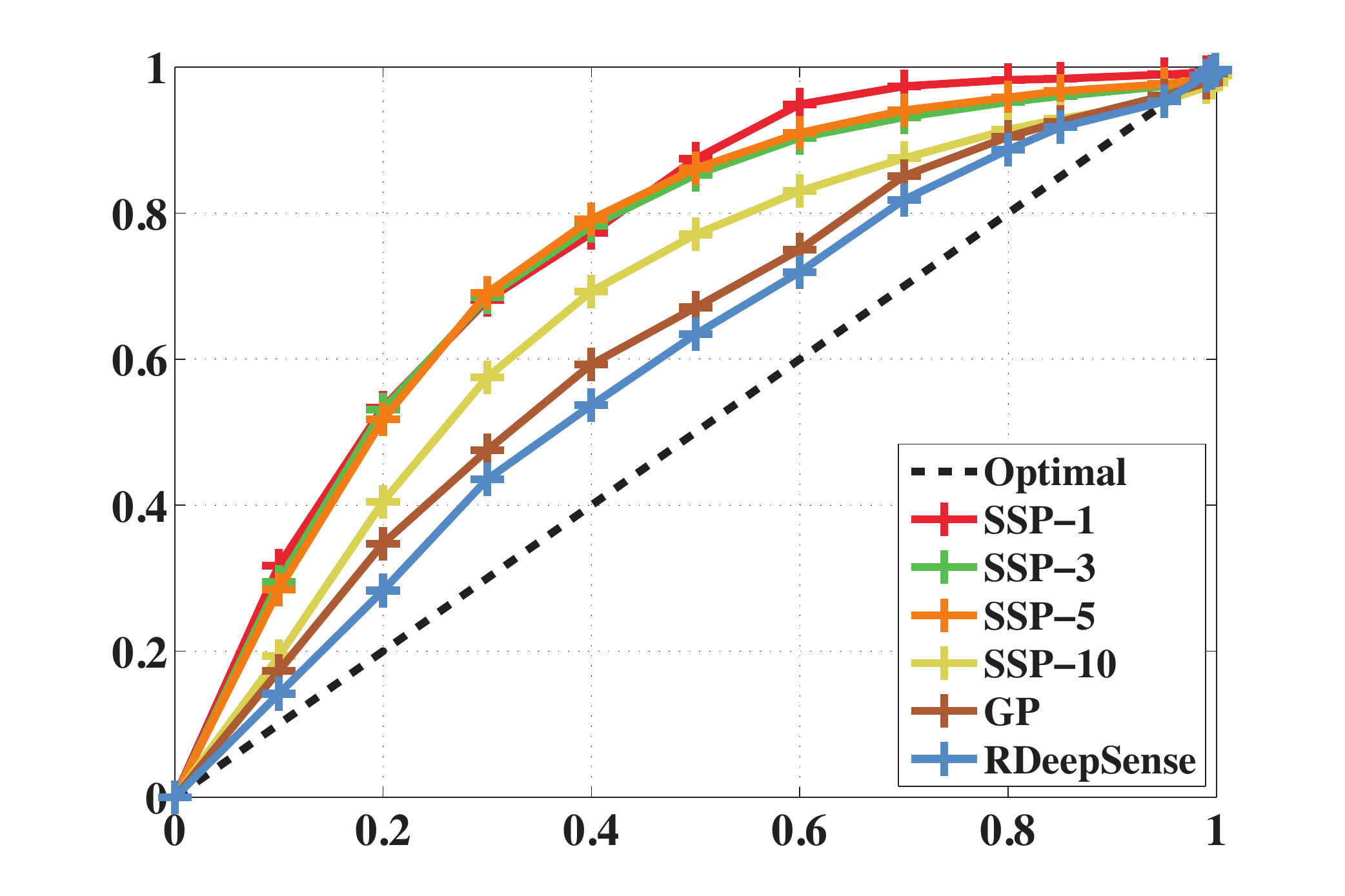}
  \caption{The calibration curves of RDeepSense, GP, and SSP-k.}
  \label{fig:GasSen_conf_SSP}
\end{subfigure}
\begin{subfigure}{.4\linewidth}
  \centering
  \includegraphics[width=1.\linewidth]{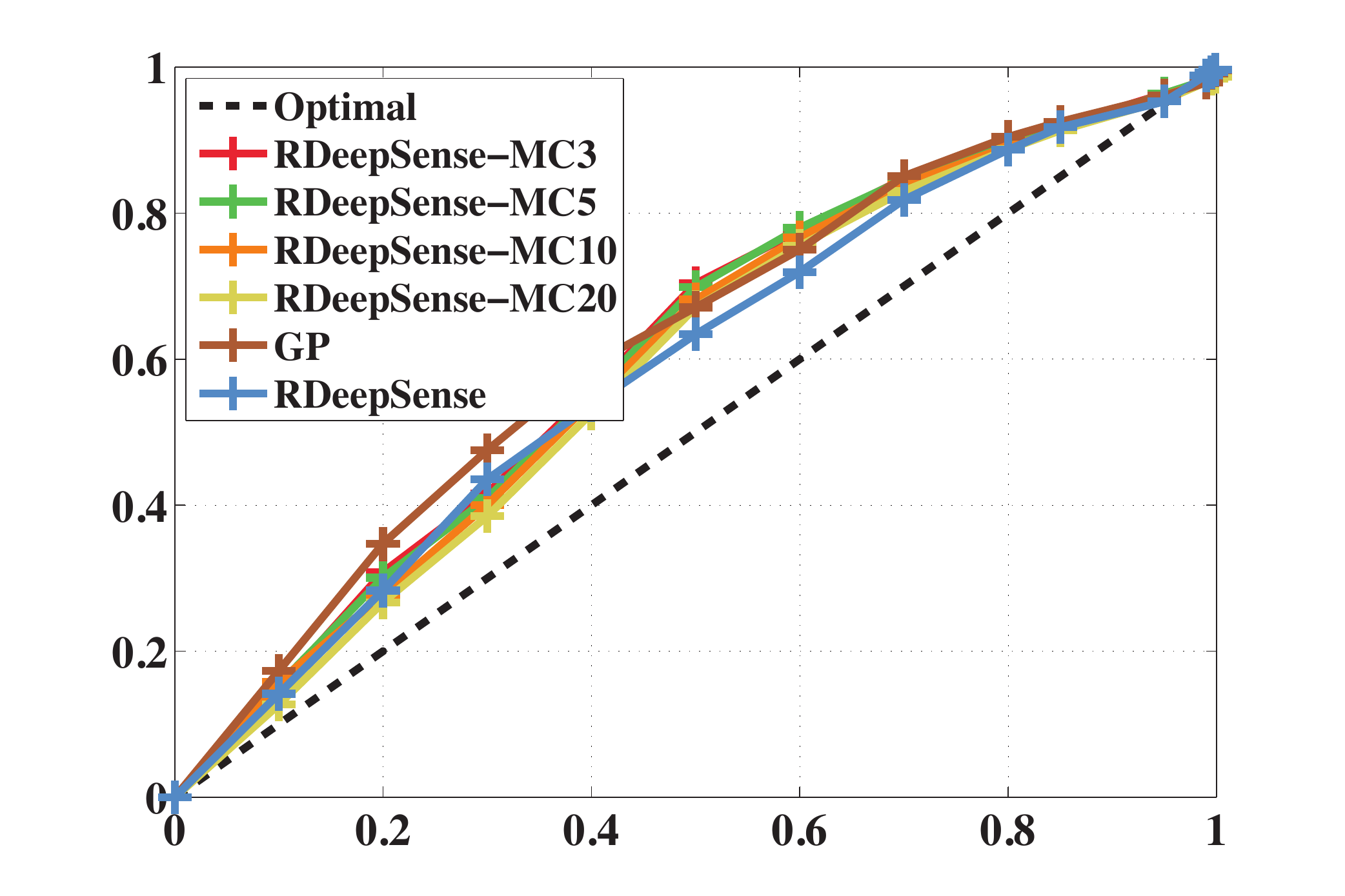}
  \caption{{\color{cmtColor}The calibration curves of RDeepSense, GP, and RDeepSense-MCk.}}
  \label{fig:GasSen_conf_RDeepSense_MC}
\end{subfigure}
\caption{{\color{cmtColor}The calibration curves of GasSen for RDeepSense, GP, MCDrop-k, SSP-k, and RDeepSense-MCk. MCDrop-k highly underestimates the predictive distribution. SSP-k highly overestimates the predictive distribution. RDeepSense is the closest curve to the optimal predictive distribution.}}
\label{fig:GasSen_conf}
%\vspace{-0.15cm}
\end{figure}

The calibration curves of GasSen task is illustrated in Figure~\ref{fig:GasSen_conf}. The calibration curves of MCDrop-k highly underestimates the predictive distribution as shown in Figure~\ref{fig:GasSen_conf_MCDrop}, while the calibration curves of SSP-k highly overestimates the predictive distribution as shown in Figure~\ref{fig:GasSen_conf_SSP}. Although there exists a bit deviation for RDeepSense compared with the optimal calibration curve, RDeepSense greatly reduces the effect of underestimation and overestimation, and slightly outperforms the traditional statistical model, GP. {\color{cmtColor}Compared with unbiased RDeepSense-MCk, RDeepSense shows the similar performance. However, RDeepSense saves save a great amount of energy and time consumption as we will discuss in Section~\ref{sec:eval_time_energy}.}

\subsubsection{HHAR}~\label{sec:HHAR_eval}
Last we compare RDeepSense with the other baseline algorithm for the HHAR task. Table~\ref{tab:exp_HHAR} illustrates the performance metrics of all algorithms based on Accuracy (Acc), {\color{cmtColor}F1 score (F1 Score)}, Negative Log-Likelihood (NLL), and Mean Entropy of False Predictions (MEFP).

\begin{table}[!htb]
%\hspace{-1.3cm}
\small
\begin{center}
\caption {{\color{cmtColor}Accuracy (Acc), Negative Log-Likelihood (NLL), Mean Entropy of False Predictions (MEFP) for the HHAR task. RDeepSense is the best-performing algorithm according to all measures.}}
\vspace{-0.1cm}
\label{tab:exp_HHAR}
\begin{tabular}{ |c | c | c | c | c | c| } 
 \hline
 & RDeepSense  & {\color{cmtColor}RDeepSense-MC3} & {\color{cmtColor}RDeepSense-MC5}  &  {\color{cmtColor}RDeepSense-MC10} & {\color{cmtColor}RDeepSense-MC20}  \\
 \hline
  Acc & $\mathbf{83.98\%}$ & $80.66\%$ & $83.07\%$ & $83.08\%$ & $83.85\%$  \\ 
  \hline
  {\color{cmtColor}F1 Score} & $\mathbf{0.670}$ & $0.601$ & $0.638$ & $0.668$  &  $\mathbf{0.671}$  \\ 
 \hline
  NLL & $\mathbf{0.161}$ & $0.193$ & $0.188$ & $0.172$  &  $\mathbf{0.159}$  \\ 
 \hline
  MEFP & $\mathbf{1.715}$ & $1.604$ & $1.621$ & $1.626$  &  $1.628$  \\ 
 \hline
  \hline
 & SSP-1  & SSP-3 & SSP-5  &  SSP-10 & GP \\
  \hline
  Acc &  $77.15\%$ & $78.34\%$ &$79.30\%$ & $80.30\%$ & $77.29\%$    \\ 
  \hline
   {\color{cmtColor}F1 Score} & $0.650$ & $0.652$ & $0.657$ & $0.661$  &  $0.659$  \\ 
 \hline
  NLL &  $1.138$ & $1.188$ & $1.165$ & $1.214$ & $0.807$   \\ 
 \hline
  MEFP &  $1.619$ & $1.629$ & $1.672$ & $1.708$ & $1.218$    \\ 
 \hline
 \hline
 & MCDrop-3  & MCDrop-5 & MCDrop-10  &  MCDrop-20 &  \\
  \hline
   Acc & $79.53\%$ & $79.73\%$& $79.73\%$ & $80.51\%$ &  \\ 
  \hline
  {\color{cmtColor}F1 Score} & $0.586$ & $0.589$ & $0.589$ & $0.593$  &  \\ 
 \hline
  NLL & $0.166$ & $0.163$ & $0.162$ & $\mathbf{0.161}$  &  \\ 
 \hline
  MEFP & $0.501$ & $0.548$ & $0.574$ & $0.579$  &  \\ 
 \hline
\end{tabular}
\end{center}
%\vspace{-0.6cm}
\end{table}

Except for RDeepSense-MC20, RDeepSense is the best-performing algorithm according to all measures, which means RDeepSense can provide both high prediction accuracy as well as high quality of uncertainty estimations. MCDrop-k algorithms are trained with log-likelihood. Therefore they try to minimize the negative log-likelihood, but they are over-confident about their prediction even when they make some wrong predictions according to the MEFP measure. SSP-k algorithms are trained with Brier score. Therefore they fall short to achieve smaller NLL values. {\color{cmtColor}Compared with RDeepSense-MCk algorithms, RDeepSense still provides a good performance in all measurements. Only RDeepSense-MC20 shows a superior performance on F1 score and NLL measurements.}

\subsection{Inference time and energy consumption}~\label{sec:eval_time_energy}

We compared the resource consumption of each algorithm including inference time and energy consumption of one-data-sample execution, which are two key issues for mobile and ubiquitous computing. All the experiments are conducted on Intel Edison with only CPU as the computing unit. No further optimization is made on any algorithms. The inference time and energy consumption of GP are not included. This is because the time complexity of GP is $O(N^3)$, where $N$ is the size of training dataset, which is infeasible for embedded devices such as Intel Edison.
The results of four tasks, \ie BPEst, NYCommute, GasSen, and HHAR, are illustrated in Figures~\ref{fig:BPEst_time_energy}, \ref{fig:NYCommute_time_energy}, \ref{fig:GasSen_time_energy}, and \ref{fig:HHAR_time_energy} respectively.

\begin{figure}[!htb]
%\vspace{-0.3cm}
\begin{subfigure}{.48\linewidth}
  \centering
  \includegraphics[width=0.85\linewidth]{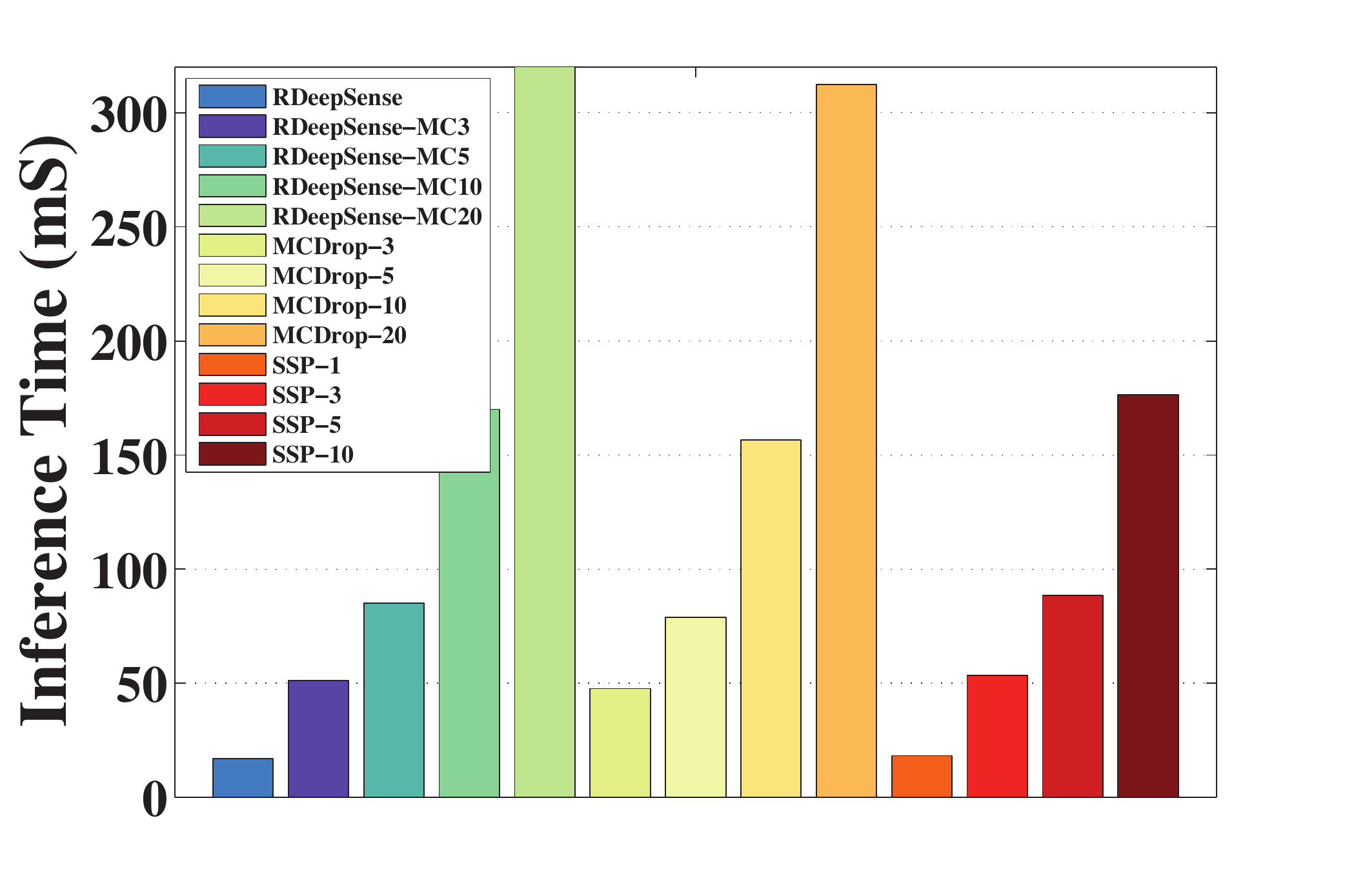}
  \caption{{\color{cmtColor}The inference time of RDeepSense, RDeepSense-MCk, MCDrop-k, and SSP-k for BPEst.}}
  \label{fig:BPEst_time}
\end{subfigure}
\begin{subfigure}{.48\linewidth}
  \centering
  \includegraphics[width=0.85\linewidth]{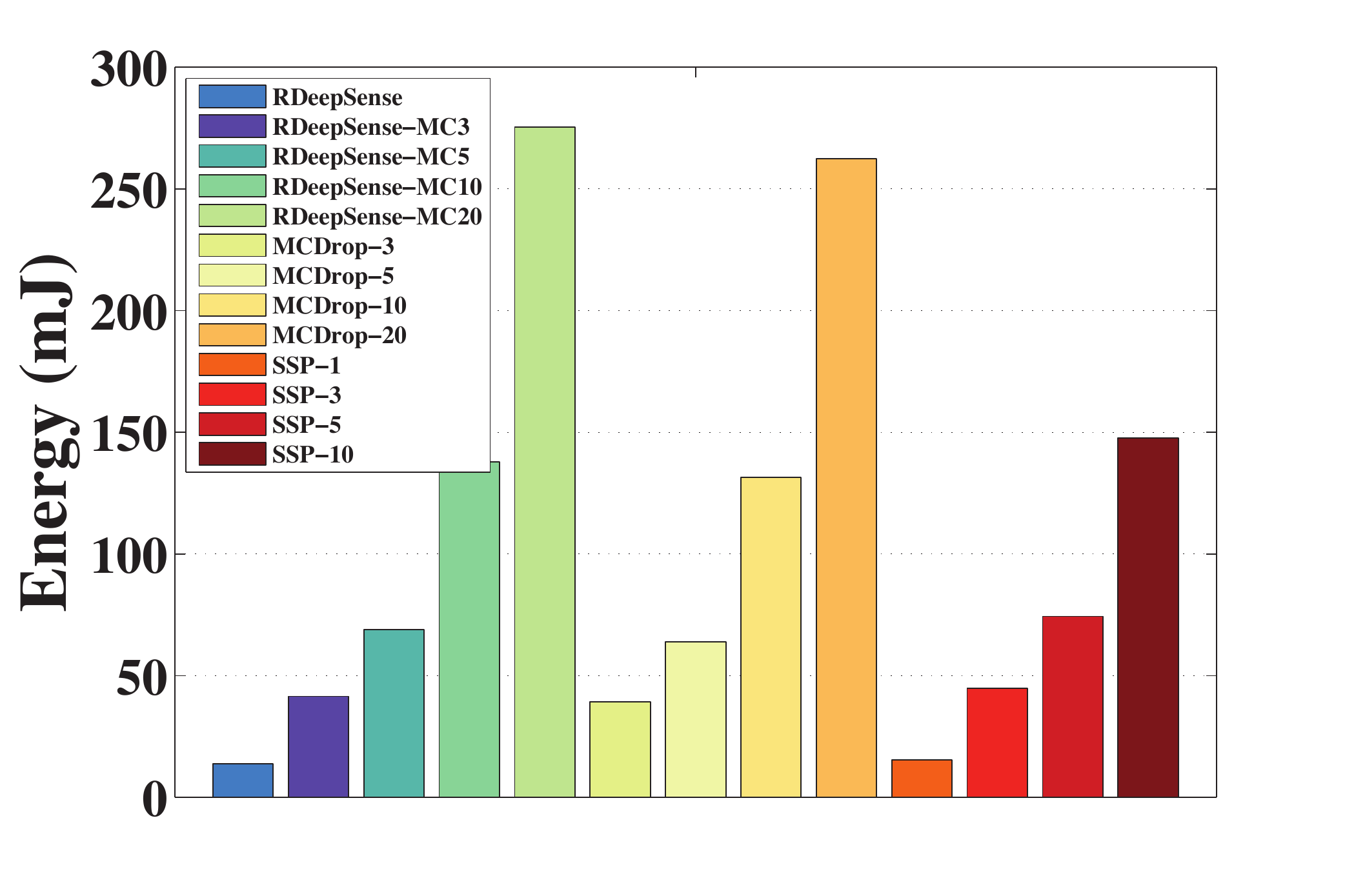}
  \caption{{\color{cmtColor}The energy consumption of RDeepSense, RDeepSense-MCk,MCDrop-k, and SSP-k for BPEst.}}
  \label{fig:BPEst_energy}
\end{subfigure}
\caption{{\color{cmtColor}The inference time and energy consumption of RDeepSense, RDeepSense-MCk, MCDrop-k, and SSP-k for BPEst.}}
\label{fig:BPEst_time_energy}
\vspace{-0.3cm}
\end{figure}

\begin{figure}[!htb]
%\vspace{-0.3cm}
\begin{subfigure}{.48\linewidth}
  \centering
  \includegraphics[width=0.85\linewidth]{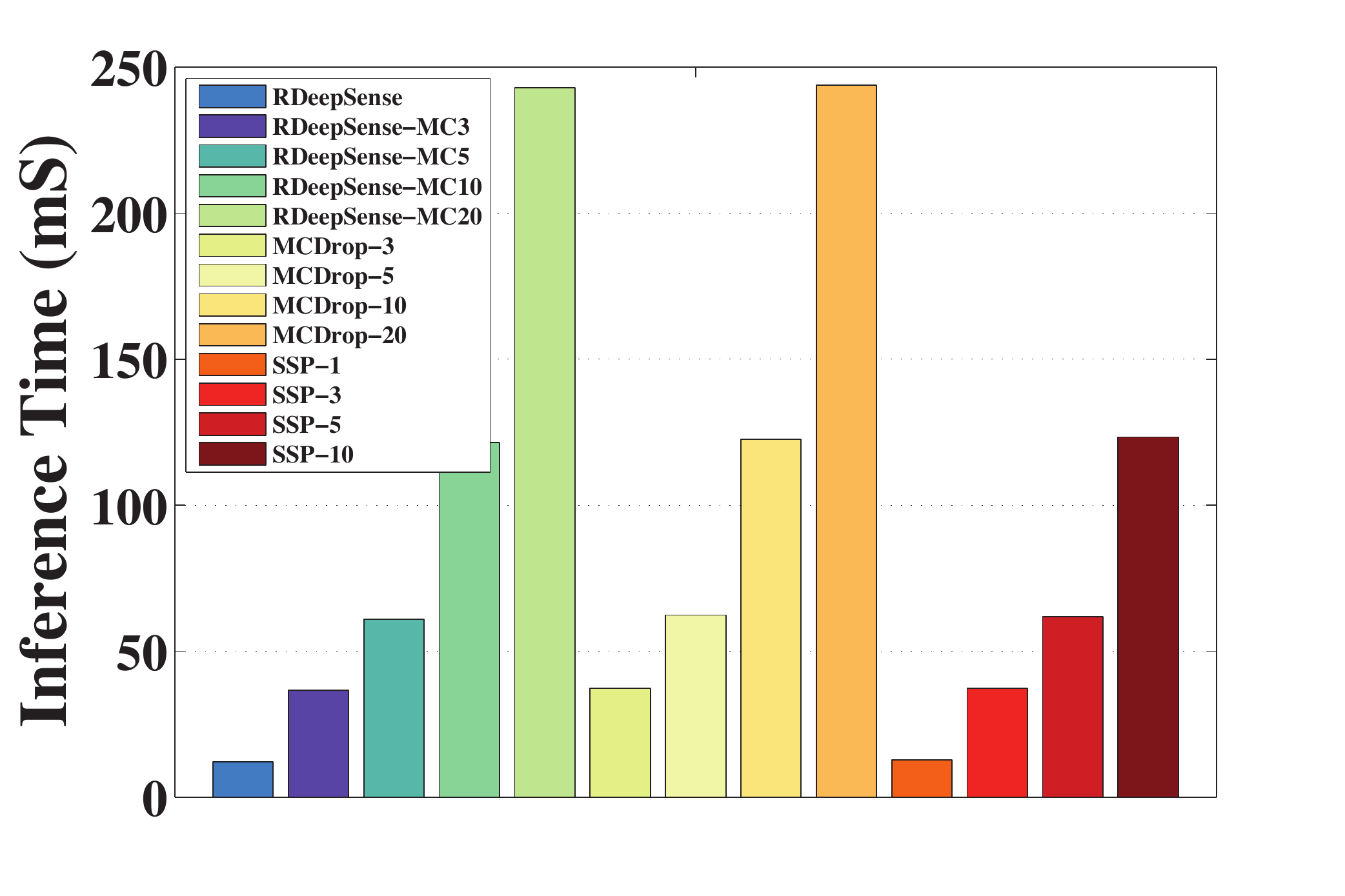}
  \caption{{\color{cmtColor}The inference time of RDeepSense, RDeepSense-MCk, MCDrop-k, and SSP-k for NYCommute.}}
  \label{fig:NYCommute_time}
\end{subfigure}
\begin{subfigure}{.48\linewidth}
  \centering
  \includegraphics[width=0.85\linewidth]{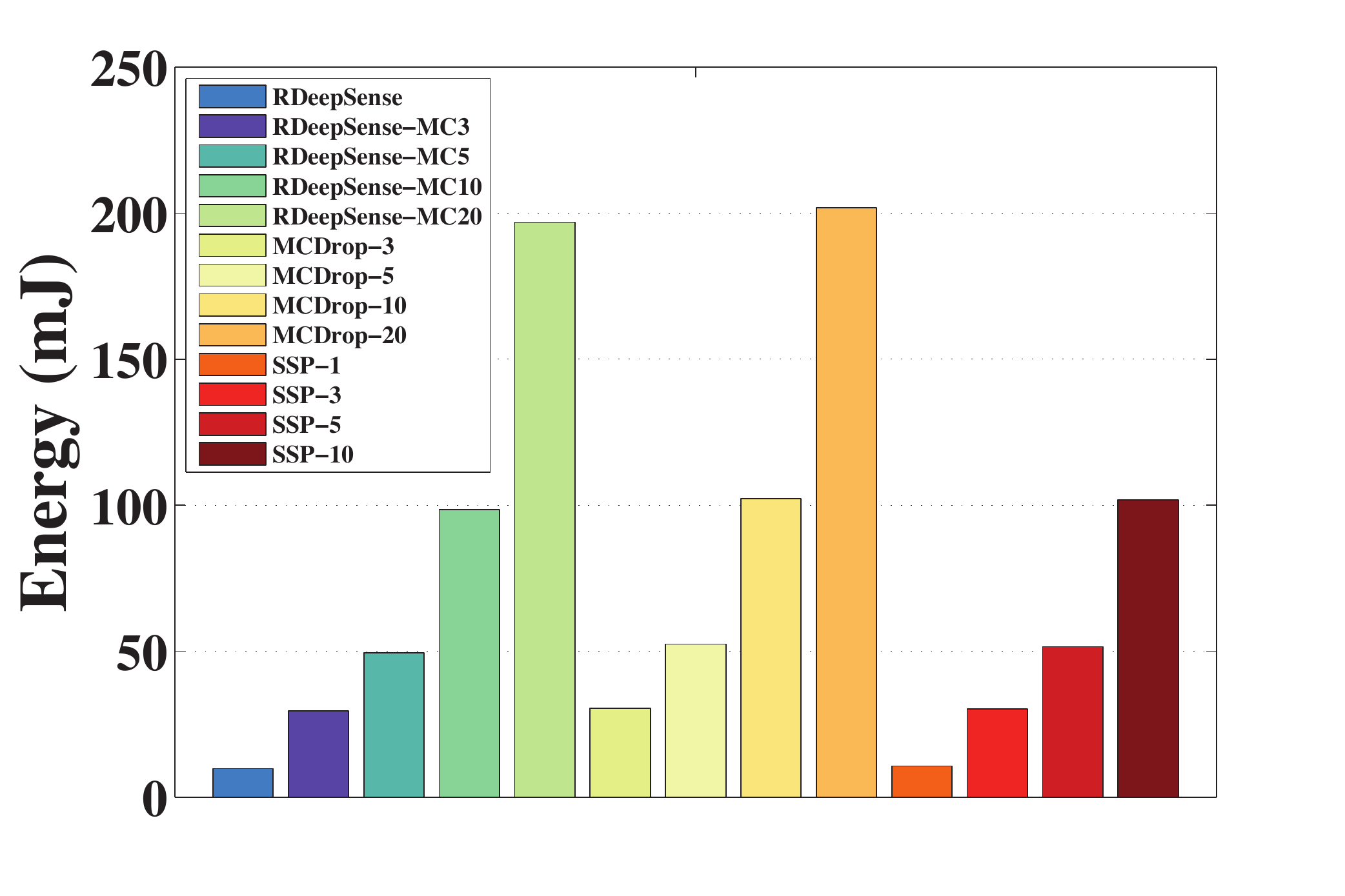}
  \caption{{\color{cmtColor}The energy consumption of RDeepSense, RDeepSense-MCk, MCDrop-k, and SSP-k for NYCommute.}}
  \label{fig:NYCommute_energy}
\end{subfigure}
\caption{{\color{cmtColor}The inference time and energy consumption of RDeepSense, RDeepSense-MCk, MCDrop-k, and SSP-k for NYCommute.}}
\label{fig:NYCommute_time_energy}
\vspace{-0.3cm}
\end{figure}

\begin{figure}[!htb]
%\vspace{-0.3cm}
\begin{subfigure}{.48\linewidth}
  \centering
  \includegraphics[width=0.85\linewidth]{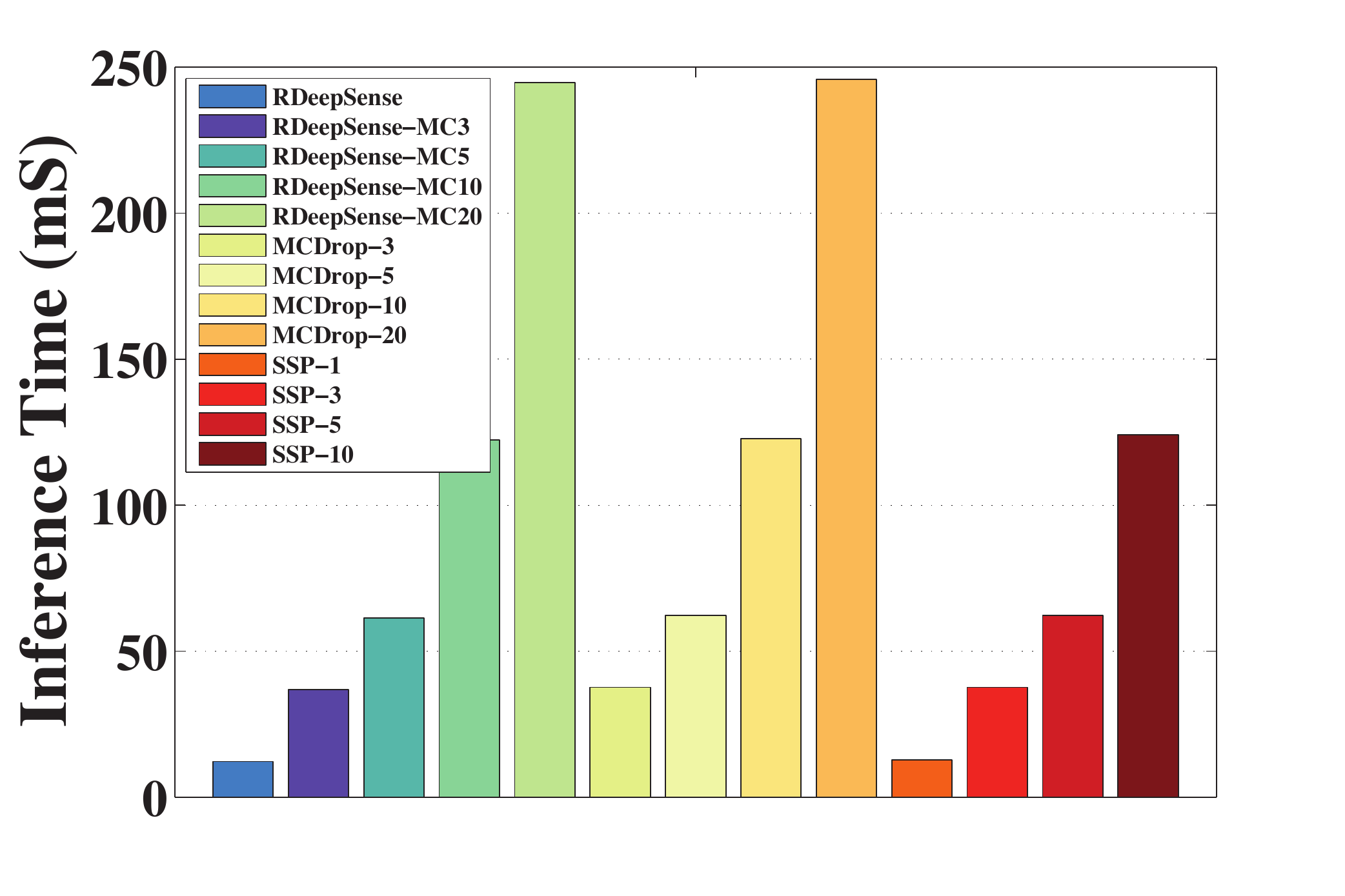}
  \caption{{\color{cmtColor}The inference time of RDeepSense, RDeepSense-MCk, MCDrop-k, and SSP-k for GasSen.}}
  \label{fig:GasSen_time}
\end{subfigure}
\begin{subfigure}{.48\linewidth}
  \centering
  \includegraphics[width=0.85\linewidth]{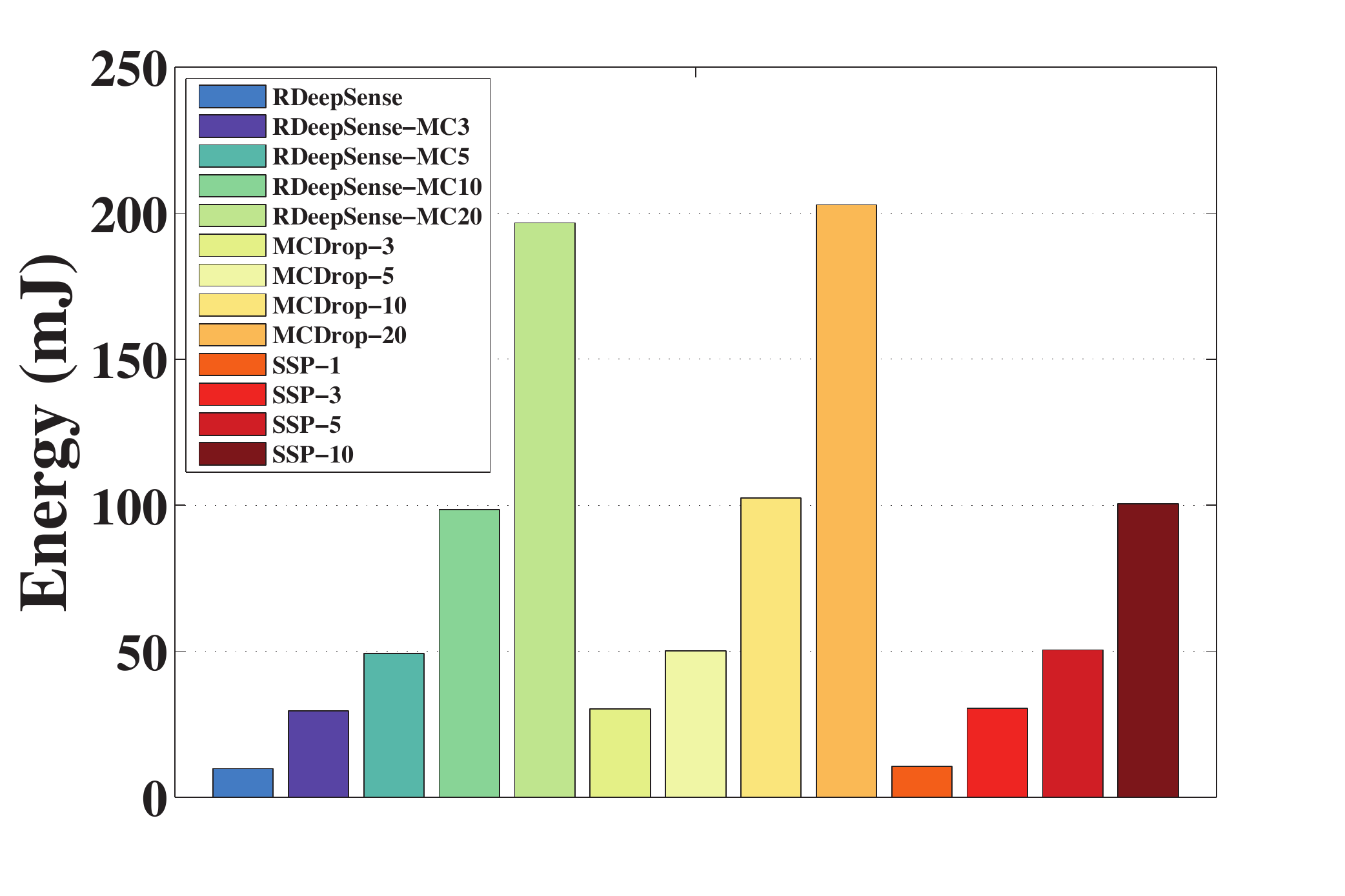}
  \caption{{\color{cmtColor}The energy consumption of RDeepSense, RDeepSense-MCk, MCDrop-k, and SSP-k for GasSen.}}
  \label{fig:GasSen_energy}
\end{subfigure}
\caption{{\color{cmtColor}The inference time and energy consumption of RDeepSense, RDeepSense-MCk, MCDrop-k, and SSP-k for GasSen.}}
\label{fig:GasSen_time_energy}
\vspace{-0.3cm}
\end{figure}

\begin{figure}[!htb]
%\vspace{-0.3cm}
\begin{subfigure}{.48\linewidth}
  \centering
  \includegraphics[width=0.85\linewidth]{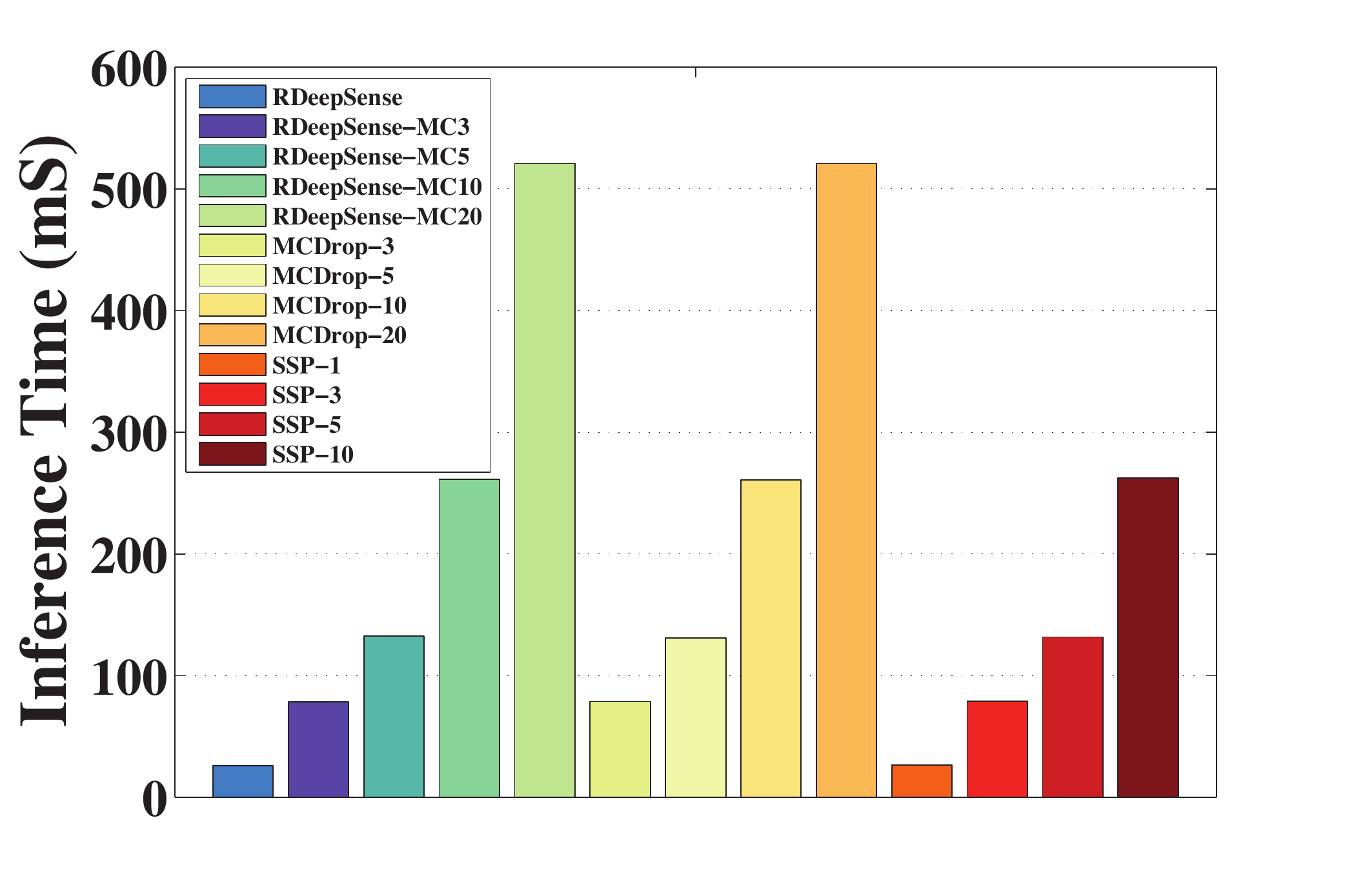}
  \caption{{\color{cmtColor}The inference time of RDeepSense, RDeepSense-MCk, MCDrop-k, and SSP-k for HHAR.}}
  \label{fig:HHAR_time}
\end{subfigure}
\begin{subfigure}{.48\linewidth}
  \centering
  \includegraphics[width=0.85\linewidth]{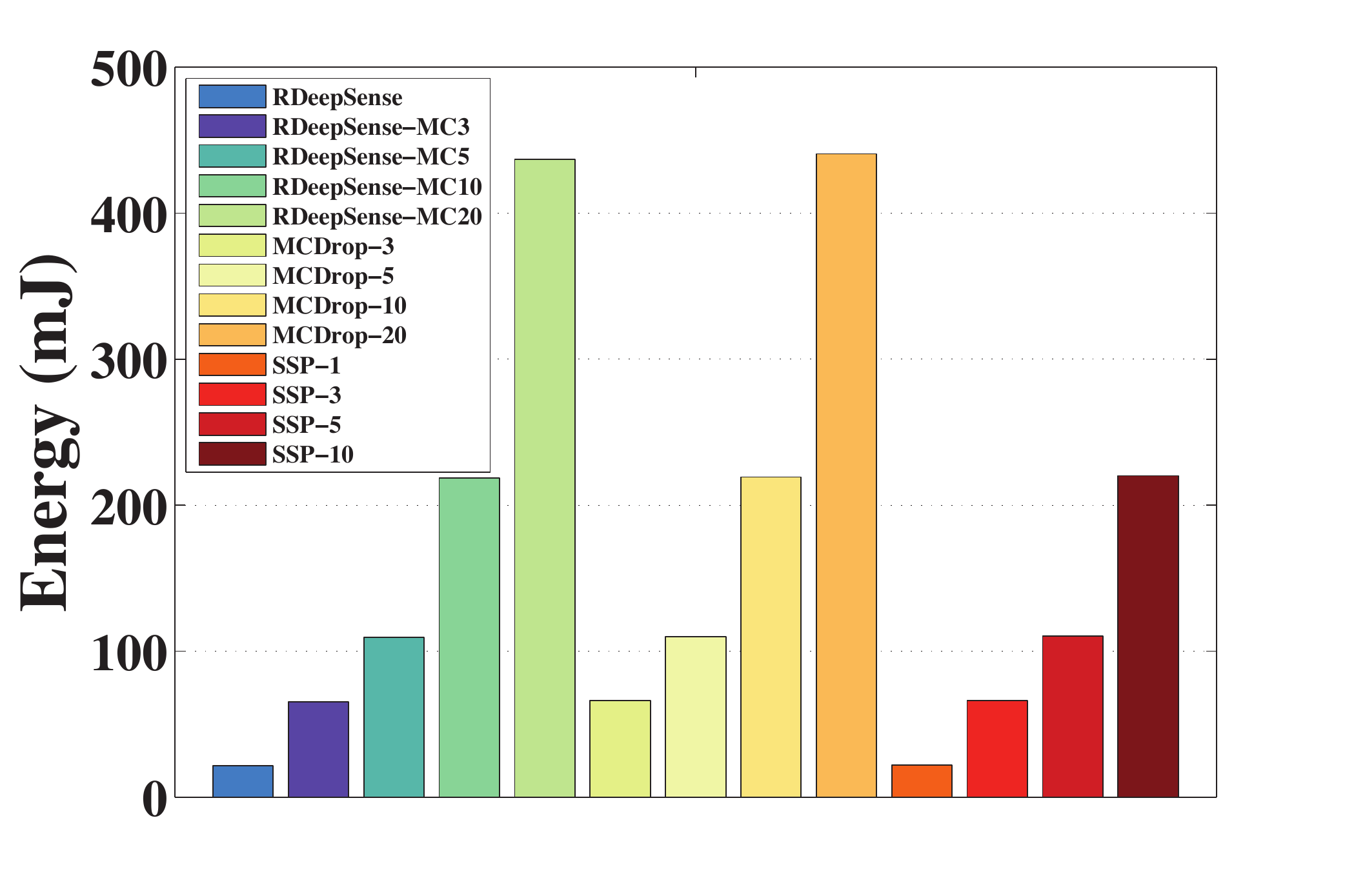}
  \caption{{\color{cmtColor}The energy consumption of RDeepSense, RDeepSense-MCk, MCDrop-k, and SSP-k for HHAR.}}
  \label{fig:HHAR_energy}
\end{subfigure}
\caption{{\color{cmtColor}The inference time and energy consumption of RDeepSense, RDeepSense-MCk, MCDrop-k, and SSP-k for HHAR.}}
\label{fig:HHAR_time_energy}
\vspace{-0.15cm}
\end{figure}

\begin{figure}[!htb]
%\vspace{-0.3cm}
\begin{subfigure}{.4\linewidth}
  \centering
  \includegraphics[width=1.\linewidth]{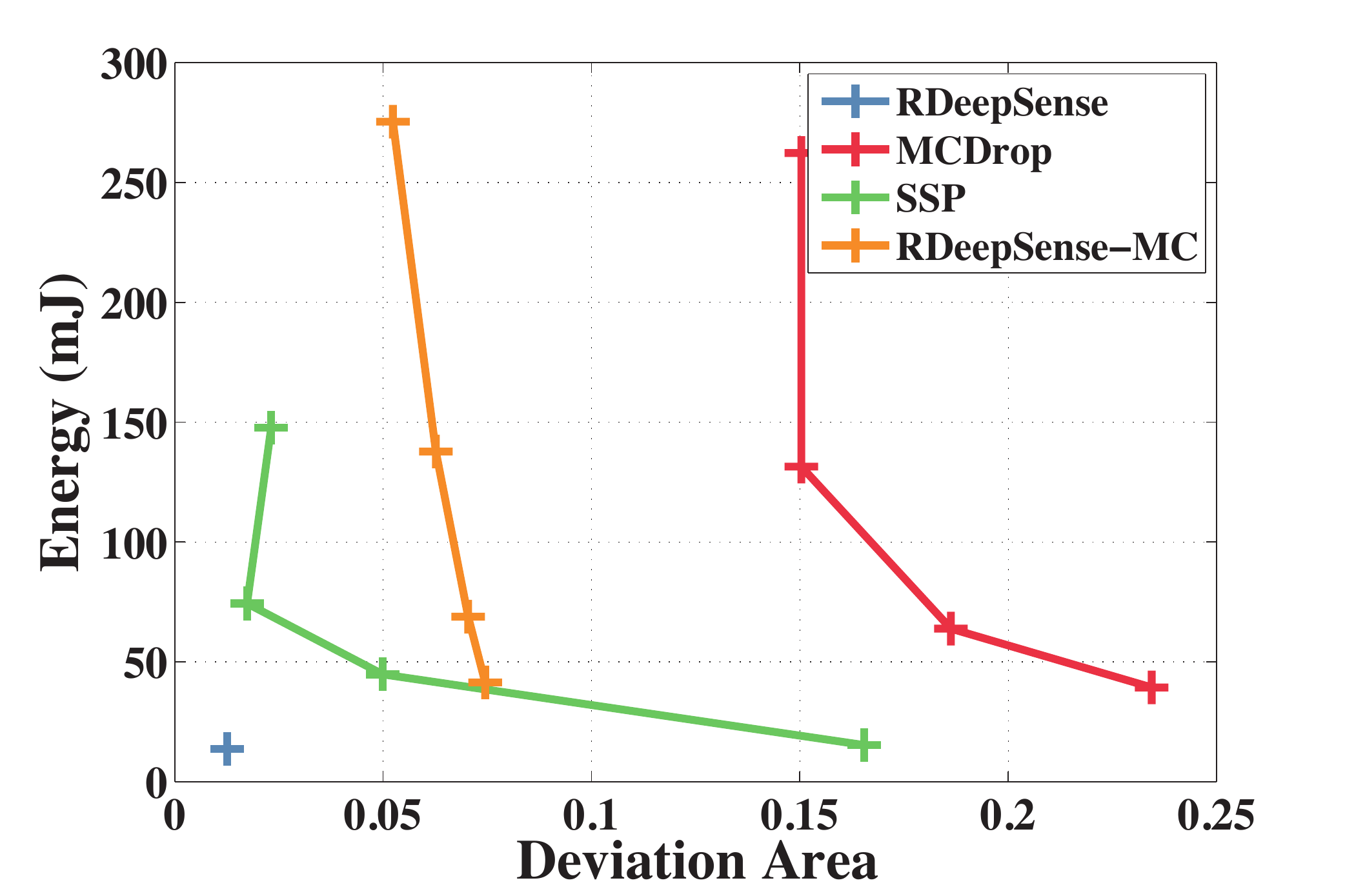}
  \caption{{\color{cmtColor}The relationship between deviation area and energy consumption for BPEst.}}
  \label{fig:qaEng_BPEst}
\end{subfigure}
\begin{subfigure}{.4\linewidth}
  \centering
  \includegraphics[width=1.\linewidth]{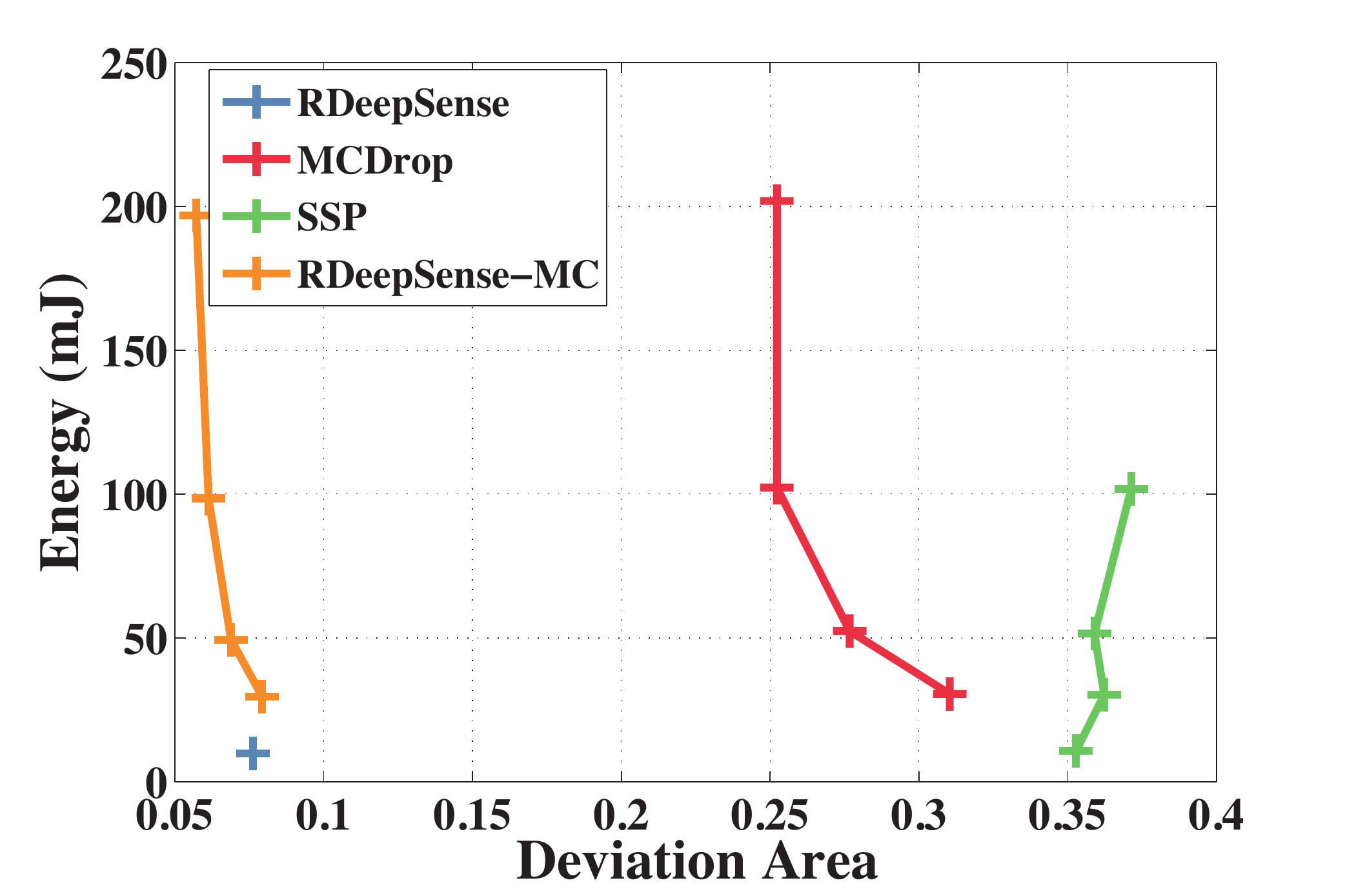}
  \caption{{\color{cmtColor}The relationship between deviation area and energy consumption for NYCommute.}}
  \label{fig:qaEng_NYCommute}
\end{subfigure}
\begin{subfigure}{.4\linewidth}
  \centering
  \includegraphics[width=1.\linewidth]{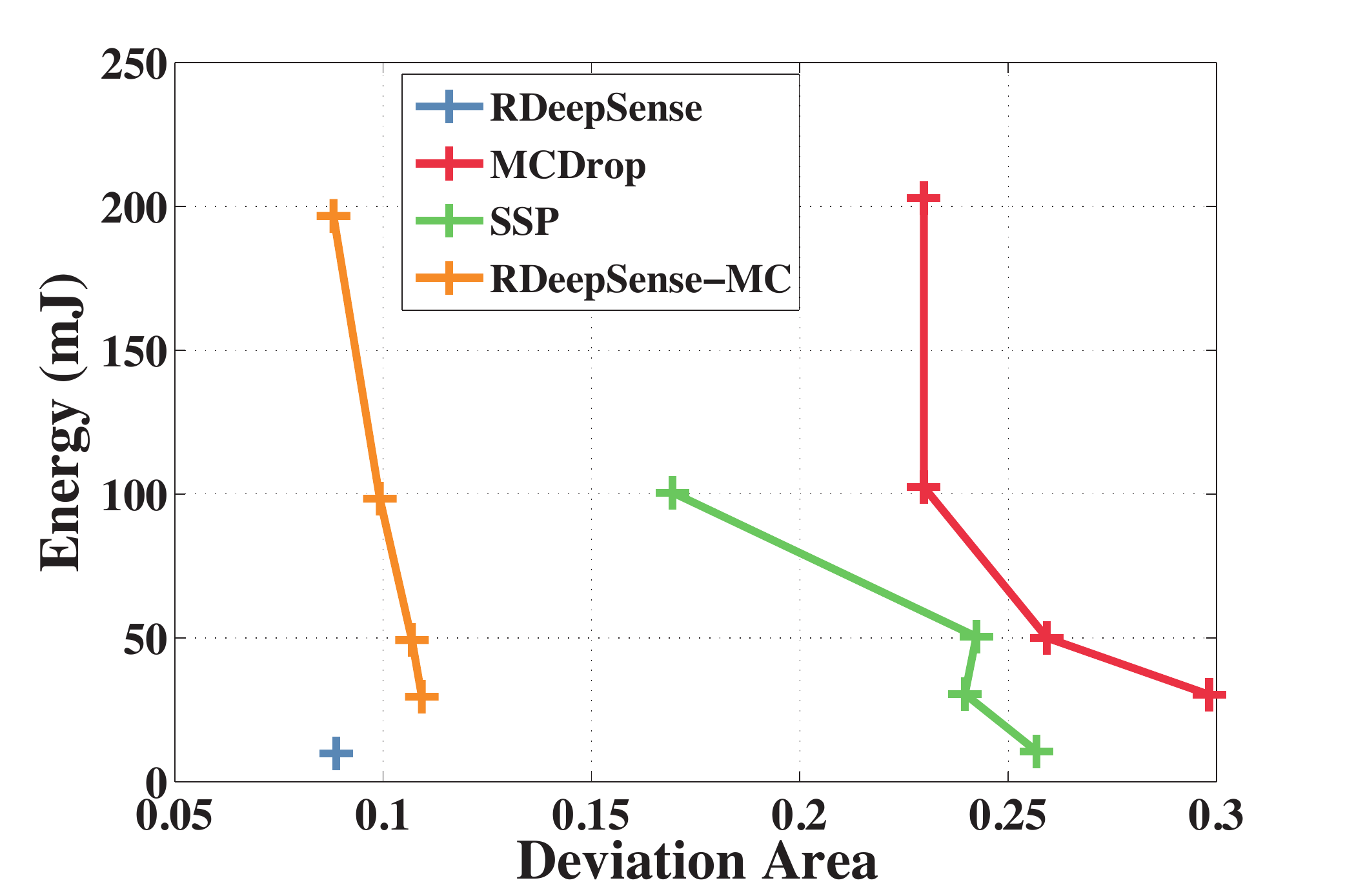}
  \caption{{\color{cmtColor}The relationship between deviation area and energy consumption for GasSen.}}
  \label{fig:qaEng_GasSen}
\end{subfigure}
\begin{subfigure}{.4\linewidth}
  \centering
  \includegraphics[width=1.\linewidth]{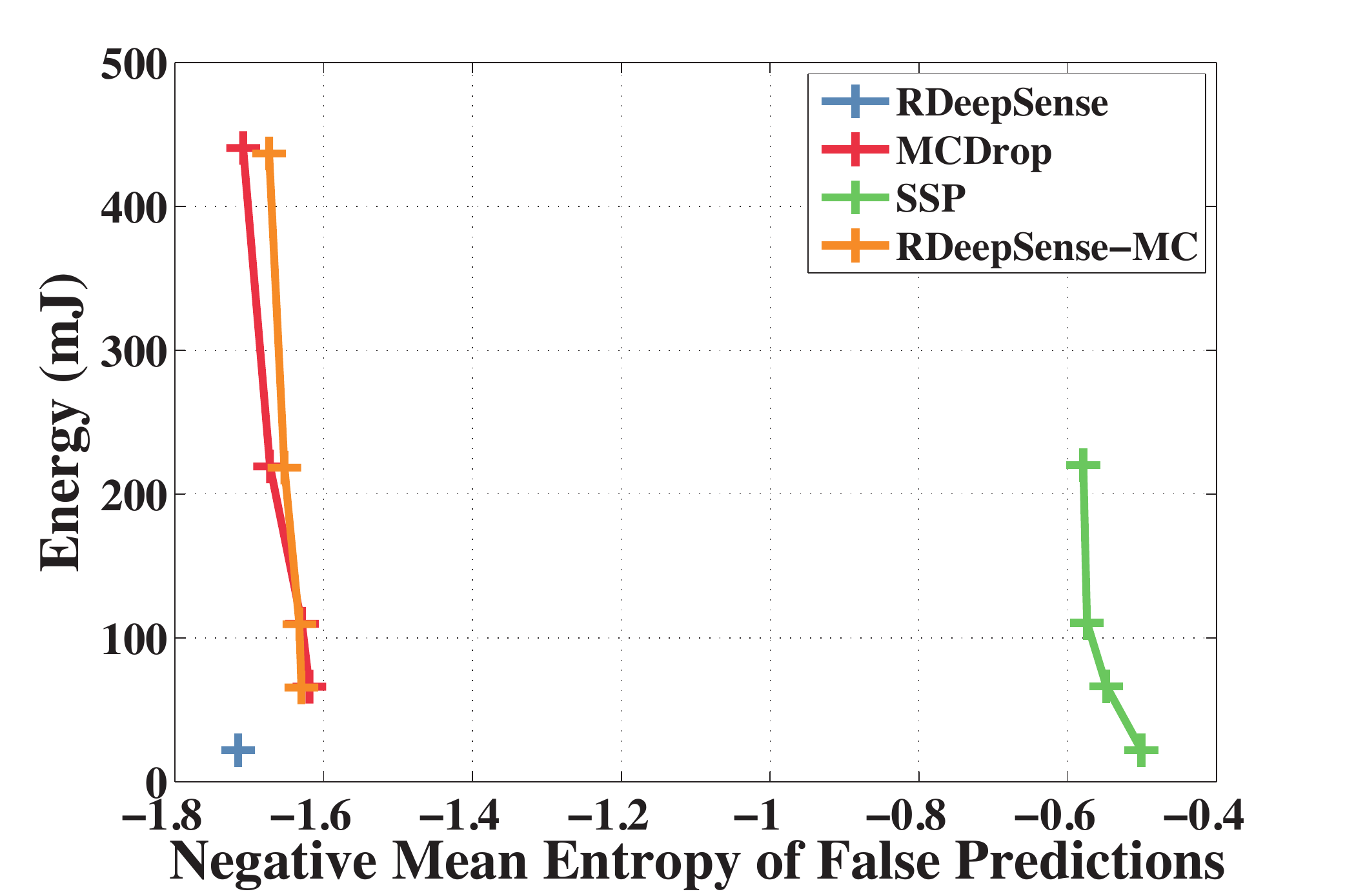}
  \caption{{\color{cmtColor}The relationship between negative mean entropy of false predictions and energy consumption for HHAR.}}
  \label{fig:qaEng_HHAR}
\end{subfigure}
\caption{{\color{cmtColor}The relationship between deviation area/negative mean entropy of false predictions and energy consumption of all algorithms. RDeepSense (in the bottom-left corner) is the best-performing algorithm that uses the least energy to achieve the best uncertainty estimation quality}}
\label{fig:qaEng}
%\vspace{-0.15cm}
\end{figure}

We can clearly see that RDeepSense greatly reduces the inference time and energy consumption compared with the other deep learning uncertainty estimation algorithms. Compared with MCDrop algorithm, RDeepSense is trained according to the proper scoring rule, which can directly output the predictive distribution instead of using sampling methods. Compared with SSP algorithm, RDeepSense uses dropout regularization as an implicit ensemble method, which avoids running multiple deep learning models during model inference on embedded devices. {\color{cmtColor}Compared with RDeepSense-MC, RDeepSense use the approximation~\eqref{eqn:dropout_test} to replace the computationally intensive Motel Carlo method~\eqref{eqn:pred_dist_reg} during the inference.}

We further analyze the relationship between energy consumption and the quality of uncertainty estimation for each algorithms. For regression problems, we use the area between the calibration curve of an algorithm and the optimal calibration curve, called deviation area, as the quality measurement of uncertainty. The smaller deviation area is, the better quality of uncertainty the algorithm estimates. When the calibration curve of an algorithm is optimal, the deviation area is $0$. For classification problems, we use the negative mean entropy of false predictions as the quality measurement of uncertainty. Smaller negative mean entropy of false predictions means is that the algorithm is more uncertain about their false predictions. The result is shown in Figure~\ref{fig:qaEng}.

The point or line stay in the bottom-left corner of the graph represents a better tradeoff between energy and uncertainty quality, \ie using less energy to obtain better uncertainty estimations. Therefore, RDeepSense is the best-performing algorithm that uses the least amount of energy to obtain the best uncertainty estimation quality. 
{\color{cmtColor}RDeepSense-MC can achieve similar uncertainty estimation quality as RDeepSense, however it requires much more energy consumption. The results show that RDeepSense is an effective and efficient uncertainty estimation algorithm~\eqref{eqn:dropout_test} compared with its Monte Carlo version~\eqref{eqn:pred_dist_reg}.}
Other two baseline algorithms, MCDrop and SSP,  usually suffer a large deviation area or 
become over-confidence about their false predictions while using more energy for computation. 
Figure~\ref{fig:qaEng} shows that RDeepSense is the most suitable algorithm for generate predictive uncertainty estimations for mobile and ubiquitous computing application on embedded devices.

\begin{figure}[!htb]
%\vspace{-0.3cm}
\begin{subfigure}{.45\linewidth}
  \centering
  \includegraphics[width=1.\linewidth]{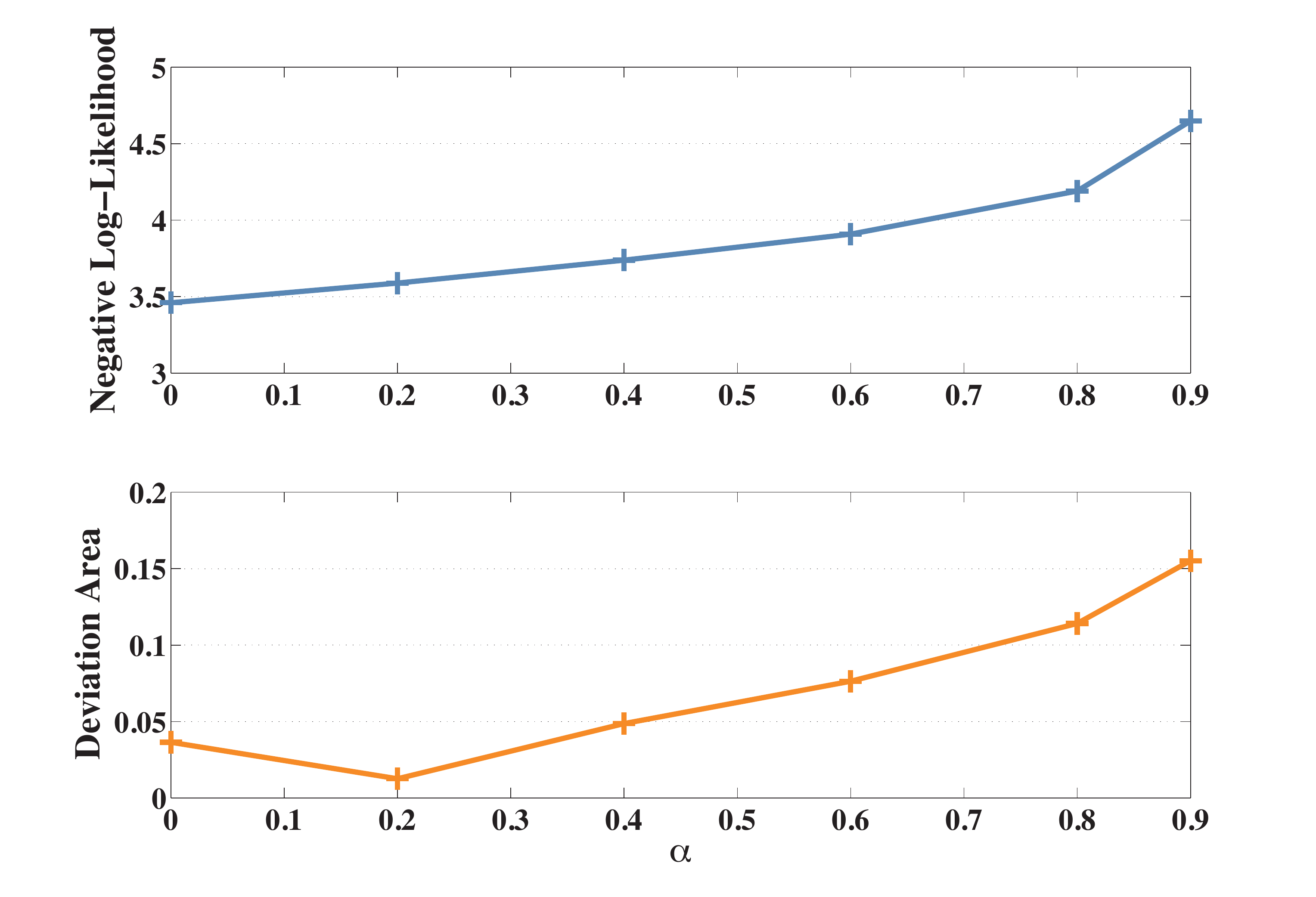}
  \caption{{\color{cmtColor}Negative Log-Likelihood and Deviation Area with different selections of $\alpha$ for BPEst.}}
  \label{fig:alpha_test_BPEst}
\end{subfigure}
\begin{subfigure}{.45\linewidth}
  \centering
  \includegraphics[width=1.\linewidth]{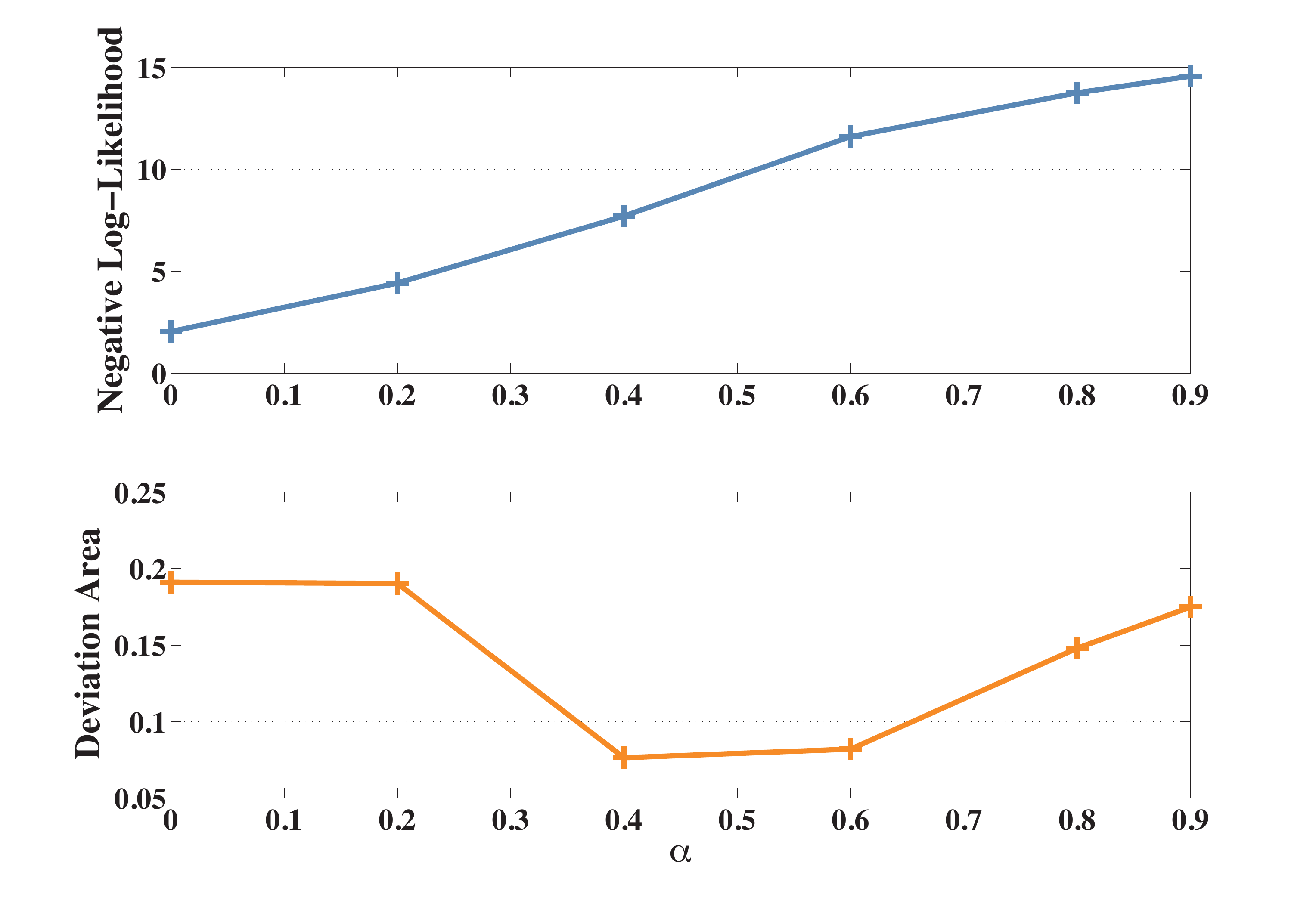}
  \caption{{\color{cmtColor}Negative Log-Likelihood and Deviation Area with different selections of $\alpha$ for NYCommute.}}
  \label{fig:alpha_test_NYCommute}
\end{subfigure}
\begin{subfigure}{.45\linewidth}
  \centering
  \includegraphics[width=1.\linewidth]{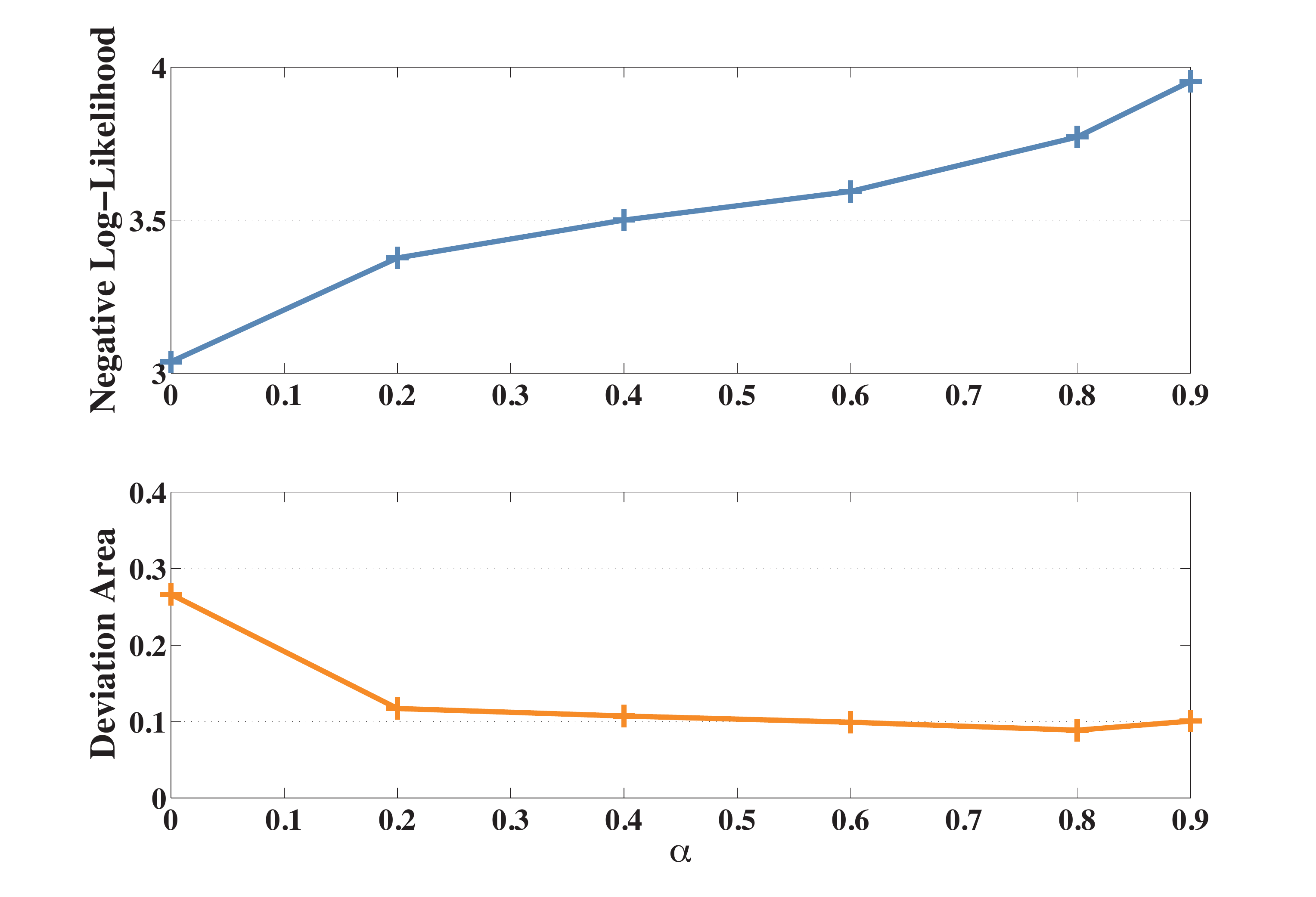}
  \caption{{\color{cmtColor}Negative Log-Likelihood and Deviation Area with different selections of $\alpha$ for GasSen.}}
  \label{fig:alpha_test_GasSen}
\end{subfigure}
\begin{subfigure}{.45\linewidth}
  \centering
  \includegraphics[width=1.\linewidth]{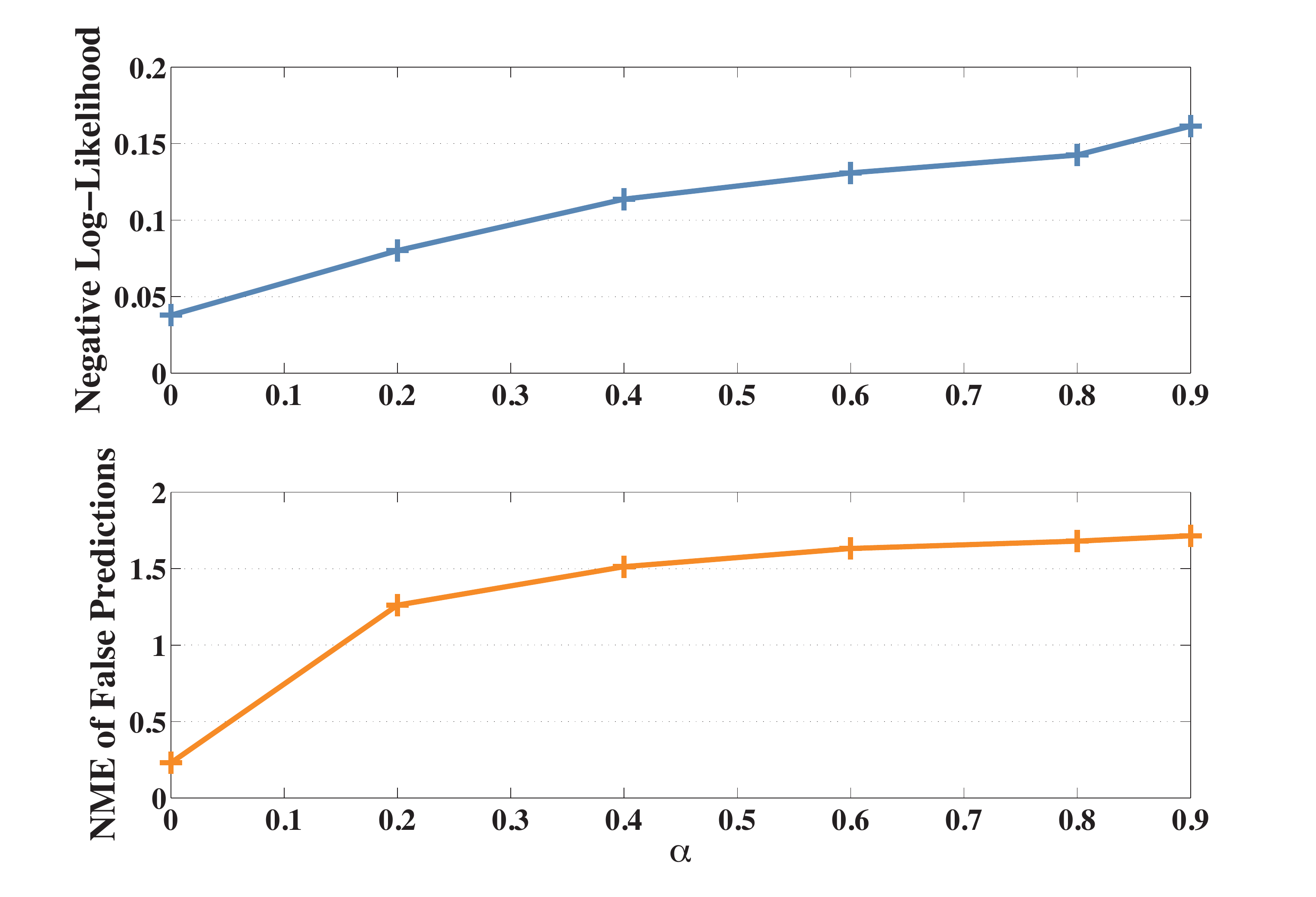}
  \caption{{\color{cmtColor}Negative Log-Likelihood and Negative Mean Entropy (NME) of false predictions with different selections of $\alpha$ for HHAR.}}
  \label{fig:alpha_test_HHAR}
\end{subfigure}
\caption{{\color{cmtColor}Negative Log-Likelihood and Deviation Area/Negative Mean Entropy (NME) of false predictions with different selections of $\alpha$ for four tasks.}}
\label{fig:alpha_test}
%\vspace{-0.15cm}
\end{figure}

{\color{cmtColor}
\subsection{Effect of hyper-parameter $\alpha$ on model performance}~\label{sec:alpha_evaluation}
The hyper-parameter $\alpha$ controls the tradeoff between optimization of mean and variance within the training objective function~\eqref{eqn:weighted_loss} that can help to obtain a well-calibrated uncertainty estimation. In this subsection, we evaluate the functionality of $\alpha$ and also shed light on the way of tuning $\alpha$.

For each task, we train RDeepSense with $\alpha = [0, 0.2, 0.4, 0.6, 0.8, 0.9]$. When $\alpha = 0.0$, RDeepSense is trained by minimizing the negative log-likelihood. When we increase the value of $\alpha$, RDeepSense focuses more on the mean value estimation instead of the negative log-likelihood. In order to show the effect of the choice of $\alpha$ on the quality of predictive uncertainty estimation, we show the negative log-likelihood and devision area (the area between the calibration curve of an algorithm and the optimal calibration curve) for regression tasks and show the negative log-likelihood and Negative Mean Entropy (NME) of false predictions for the classification task in Figure~\ref{fig:alpha_test}.

A good uncertainty estimation should faithfully reflect the probability that prediction will happen. Therefore, RDeepSense targets on a well-calibrated uncertainty estimation, such as the prediction with low devision area, in stead of the prediction with low negative log-likelihood. From Figure~\ref{fig:alpha_test_BPEst},~\ref{fig:alpha_test_NYCommute}, and~\ref{fig:alpha_test_GasSen}, we can see that hyper-parameter $\alpha$ controls the tradeoff between optimization mean and variance within the training objective function~\eqref{eqn:weighted_loss}. Smaller $\alpha$ tends to reduce negative log-likelihood by increasing the predictive variance, which tends to result the overestimation of predictive uncertainties. Larger $\alpha$ tends to reduce negative log-likelihood by predicting a better mean value, which tends to result the underestimation of predictive uncertainties. When tuning the hyper-parameter $\alpha$, we can easily found a point that achieve the smallest devision area by grid searching $\alpha$ from $0$ to $1$. At the same time, it is not surprising that increasing $\alpha$ can slightly increase the negative log-likelihood, since $\alpha=0$ represents regarding negative log-likelihood as the objective function. In addition, Figure~\ref{fig:alpha_test_HHAR} shows that increasing $\alpha$ can consistently increase the negative mean entropy of false predictions.
}

\section{Discussion}~\label{sec:discussion}
This paper focuses on empowering neural networks to generate high-quality predictive uncertainty estimations in a theoretically-grounded and energy-efficient manner for mobile and ubiquitous computing tasks. Currently, RDeepSense can only support fully-connected neural networks. It is possible to extend the two-step solution introduced in Section~\ref{sec:RDeepSense} to convolutional and recurrent neural networks by replacing the original dropout operation with convolutional dropout~\cite{gal2015bayesian} and recurrent dropout~\cite{gal2016theoretically}. These two dropout operations can convert convolutional neural networks and recurrent neural networks into Bayesian neural networks~\cite{gal2015bayesian,gal2016theoretically}, but additional efforts are needed to 1) theoretically prove that the extended two-step solution can equate an arbitrary neural network with a statistical model, and 2) empirically show that the extended two-step solution can provide high-quality uncertainty estimations on the real datasets. 

Another interesting extension could be empowering existing neural networks within mobile and ubiquitous computing applications to generate predictive uncertainty estimations without additional training. A lot of neural networks have already been trained with dropout operations. As shown by Gal et al.~\cite{Gal2016DropoutAA}, although these models tend to underestimate the true uncertainties, they can provide uncertainty estimations during model inference. This can be a good solution for mobile and ubiquitous computing applications that want to obtain an indictor of predictive uncertainty instead of a high-quality predictive uncertainty estimation without retraining their neural networks. However, additional efforts are needed to bypass the Monte Carlo sampling method and provide an energy-efficient method for generating uncertainty estimations on embedded devices.

In addition, for classification problems, although traditional neural networks can also output predictive distribution on each class, which contains predictive uncertainties, RDeepSense provides a high-quality predictive distribution on each class and has been proved to be equivalent to a statistical model.

\section{Conclusion}~\label{sec:conclusion}
We introduced RDeepSense, a simple yet effective solution that empowers fully-connected neural networks to generate well-calibrated predictive uncertainty estimations during model inference. RDeepSense is a computationally efficient algorithm that can provide predictive uncertainty estimations in mobile and ubiquitous computing applications with almost no additional overhead. Theoretical analysis also shows the equivalence between RDeepSense and a statistical model. We evaluated RDeepSense on four mobile and ubiquitous computing tasks, where RDeepSense outperformed the state-of-the-art baselines by significant margins on the quality of uncertainty estimations while still consuming the least amount of energy on embedded devices. In summary, RDeepSense is a simple, effective, and efficient solution for mobile and ubiquitous applications to build reliable neural networks with uncertainty estimations.

}

\bibliographystyle{abbrv}
\bibliography{reference}

\begin{thebibliography}{10}

\bibitem{Edison}
Intel edison compute module.
\newblock
  \url{http://www.intel.com/content/dam/support/us/en/documents/edison/sb/edison-module_HG_331189.pdf}.

\bibitem{baldi2013understanding}
P.~Baldi and P.~J. Sadowski.
\newblock Understanding dropout.
\newblock In {\em Advances in Neural Information Processing Systems}, pages
  2814--2822, 2013.

\bibitem{bauer2012shuteye}
J.~S. Bauer, S.~Consolvo, B.~Greenstein, J.~Schooler, E.~Wu, N.~F. Watson, and
  J.~Kientz.
\newblock Shuteye: encouraging awareness of healthy sleep recommendations with
  a mobile, peripheral display.
\newblock In {\em Proceedings of the SIGCHI Conference on Human Factors in
  Computing Systems}, pages 1401--1410. ACM, 2012.

\bibitem{baumann2016quantifying}
P.~Baumann, M.~Langheinrich, A.~Dey, and S.~Santini.
\newblock Quantifying the uncertainty of next-place predictions.
\newblock In {\em Proceedings of the 8th EAI International Conference on Mobile
  Computing, Applications and Services}, pages 74--85. ICST (Institute for
  Computer Sciences, Social-Informatics and Telecommunications Engineering),
  2016.

\bibitem{bentley2015reducing}
F.~R. Bentley, Y.-Y. Chen, and C.~Holz.
\newblock Reducing the stress of coordination: sharing travel time information
  between contacts on mobile phones.
\newblock In {\em Proceedings of the 33rd Annual ACM Conference on Human
  Factors in Computing Systems}, pages 967--970. ACM, 2015.

\bibitem{blei2017variational}
D.~M. Blei, A.~Kucukelbir, and J.~D. McAuliffe.
\newblock Variational inference: A review for statisticians.
\newblock {\em Journal of the American Statistical Association},
  (just-accepted), 2017.

\bibitem{boukhelifa2017data}
N.~Boukhelifa, M.-E. Perrin, S.~Huron, and J.~Eagan.
\newblock How data workers cope with uncertainty: A task characterisation
  study.
\newblock In {\em Proceedings of the 2017 CHI Conference on Human Factors in
  Computing Systems}, pages 3645--3656. ACM, 2017.

\bibitem{chen2014airlink}
K.-Y. Chen, D.~Ashbrook, M.~Goel, S.-H. Lee, and S.~Patel.
\newblock Airlink: sharing files between multiple devices using in-air
  gestures.
\newblock In {\em Proceedings of the 2014 ACM International Joint Conference on
  Pervasive and Ubiquitous Computing}, pages 565--569. ACM, 2014.

\bibitem{choudhury2008mobile}
T.~Choudhury, S.~Consolvo, B.~Harrison, J.~Hightower, A.~LaMarca, L.~LeGrand,
  A.~Rahimi, A.~Rea, G.~Bordello, B.~Hemingway, et~al.
\newblock The mobile sensing platform: An embedded activity recognition system.
\newblock {\em IEEE Pervasive Computing}, 7(2), 2008.

\bibitem{chung2011indoor}
J.~Chung, M.~Donahoe, C.~Schmandt, I.-J. Kim, P.~Razavai, and M.~Wiseman.
\newblock Indoor location sensing using geo-magnetism.
\newblock In {\em Proceedings of the 9th international conference on Mobile
  systems, applications, and services}, pages 141--154. ACM, 2011.

\bibitem{clyde2004model}
M.~Clyde and E.~I. George.
\newblock Model uncertainty.
\newblock {\em Statistical science}, pages 81--94, 2004.

\bibitem{damianou2013deep}
A.~Damianou and N.~Lawrence.
\newblock Deep gaussian processes.
\newblock In {\em Artificial Intelligence and Statistics}, pages 207--215,
  2013.

\bibitem{CastroHickson2015}
V.~B. E. T. G. A. H.~C. Daniel~Castro, Steven~Hickson and I.~Essa.
\newblock Predicting daily activities from egocentric images using deep
  learning.
\newblock {\em ISWC}, 2015.

\bibitem{faurholt2016behavioral}
M.~Faurholt-Jepsen, M.~Vinberg, M.~Frost, S.~Debel, E.~Margrethe~Christensen,
  J.~E. Bardram, and L.~V. Kessing.
\newblock Behavioral activities collected through smartphones and the
  association with illness activity in bipolar disorder.
\newblock {\em International journal of methods in psychiatric research},
  25(4):309--323, 2016.

\bibitem{fonollosa2015reservoir}
J.~Fonollosa, S.~Sheik, R.~Huerta, and S.~Marco.
\newblock Reservoir computing compensates slow response of chemosensor arrays
  exposed to fast varying gas concentrations in continuous monitoring.
\newblock {\em Sensors and Actuators B: Chemical}, 215:618--629, 2015.

\bibitem{gal2015bayesian}
Y.~Gal and Z.~Ghahramani.
\newblock Bayesian convolutional neural networks with bernoulli approximate
  variational inference.
\newblock {\em arXiv preprint arXiv:1506.02158}, 2015.

\bibitem{Gal2016DropoutAA}
Y.~Gal and Z.~Ghahramani.
\newblock Dropout as a bayesian approximation: Representing model uncertainty
  in deep learning.
\newblock In {\em ICML}, 2016.

\bibitem{gal2016theoretically}
Y.~Gal and Z.~Ghahramani.
\newblock A theoretically grounded application of dropout in recurrent neural
  networks.
\newblock In {\em Advances in Neural Information Processing Systems}, pages
  1019--1027, 2016.

\bibitem{ghahramani2015probabilistic}
Z.~Ghahramani.
\newblock Probabilistic machine learning and artificial intelligence.
\newblock {\em Nature}, 521(7553):452--459, 2015.

\bibitem{gneiting2007strictly}
T.~Gneiting and A.~E. Raftery.
\newblock Strictly proper scoring rules, prediction, and estimation.
\newblock {\em Journal of the American Statistical Association},
  102(477):359--378, 2007.

\bibitem{goldberger2000physiobank}
A.~L. Goldberger, L.~A. Amaral, L.~Glass, J.~M. Hausdorff, P.~C. Ivanov, R.~G.
  Mark, J.~E. Mietus, G.~B. Moody, C.-K. Peng, and H.~E. Stanley.
\newblock Physiobank, physiotoolkit, and physionet.
\newblock {\em Circulation}, 101(23):e215--e220, 2000.

\bibitem{gordon2012energy}
D.~Gordon, J.~Czerny, T.~Miyaki, and M.~Beigl.
\newblock Energy-efficient activity recognition using prediction.
\newblock In {\em Wearable Computers (ISWC), 2012 16th International Symposium
  on}, pages 29--36. IEEE, 2012.

\bibitem{griffiths2014health}
E.~Griffiths, T.~S. Saponas, and A.~Brush.
\newblock Health chair: implicitly sensing heart and respiratory rate.
\newblock In {\em Proceedings of the 2014 ACM International Joint Conference on
  Pervasive and Ubiquitous Computing}, pages 661--671. ACM, 2014.

\bibitem{grosse2016platypus}
T.~Grosse-Puppendahl, X.~Dellangnol, C.~Hatzfeld, B.~Fu, M.~Kupnik, A.~Kuijper,
  M.~R. Hastall, J.~Scott, and M.~Gruteser.
\newblock Platypus: Indoor localization and identification through sensing of
  electric potential changes in human bodies.
\newblock In {\em Proceedings of the 14th Annual International Conference on
  Mobile Systems, Applications, and Services}, pages 17--30. ACM, 2016.

\bibitem{guan2017ensembles}
Y.~Guan and T.~Ploetz.
\newblock Ensembles of deep lstm learners for activity recognition using
  wearables.
\newblock {\em arXiv preprint arXiv:1703.09370}, 2017.

\bibitem{jiang2012ariel}
Y.~Jiang, X.~Pan, K.~Li, Q.~Lv, R.~P. Dick, M.~Hannigan, and L.~Shang.
\newblock Ariel: Automatic wi-fi based room fingerprinting for indoor
  localization.
\newblock In {\em Proceedings of the 2012 ACM Conference on Ubiquitous
  Computing}, pages 441--450. ACM, 2012.

\bibitem{kachuee2015cuff}
M.~Kachuee, M.~M. Kiani, H.~Mohammadzade, and M.~Shabany.
\newblock Cuff-less high-accuracy calibration-free blood pressure estimation
  using pulse transit time.
\newblock In {\em Circuits and Systems (ISCAS), 2015 IEEE International
  Symposium on}, pages 1006--1009. IEEE, 2015.

\bibitem{kaiser2016design}
S.~Kaiser, A.~Parks, P.~Leopard, C.~Albright, J.~Carlson, M.~Goel, D.~Nassehi,
  and E.~C. Larson.
\newblock Design and learnability of vortex whistles for managing chronic lung
  function via smartphones.
\newblock In {\em Proceedings of the 2016 ACM International Joint Conference on
  Pervasive and Ubiquitous Computing}, pages 569--580. ACM, 2016.

\bibitem{kay2016ish}
M.~Kay, T.~Kola, J.~R. Hullman, and S.~A. Munson.
\newblock When (ish) is my bus?: User-centered visualizations of uncertainty in
  everyday, mobile predictive systems.
\newblock In {\em Proceedings of the 2016 CHI Conference on Human Factors in
  Computing Systems}, pages 5092--5103. ACM, 2016.

\bibitem{kay2015good}
M.~Kay, S.~N. Patel, and J.~A. Kientz.
\newblock How good is 85\%?: A survey tool to connect classifier evaluation to
  acceptability of accuracy.
\newblock In {\em Proceedings of the 33rd Annual ACM Conference on Human
  Factors in Computing Systems}, pages 347--356. ACM, 2015.

\bibitem{koehler2014indoor}
C.~Koehler, N.~Banovic, I.~Oakley, J.~Mankoff, and A.~K. Dey.
\newblock Indoor-alps: an adaptive indoor location prediction system.
\newblock In {\em Proceedings of the 2014 ACM International Joint Conference on
  Pervasive and Ubiquitous Computing}, pages 171--181. ACM, 2014.

\bibitem{koller2009probabilistic}
D.~Koller and N.~Friedman.
\newblock {\em Probabilistic graphical models: principles and techniques}.
\newblock MIT press, 2009.

\bibitem{krzywinski2013points}
M.~Krzywinski and N.~Altman.
\newblock Points of significance: importance of being uncertain.
\newblock {\em Nature methods}, 10(9):809, 2013.

\bibitem{lakshminarayanan2016simple}
B.~Lakshminarayanan, A.~Pritzel, and C.~Blundell.
\newblock Simple and scalable predictive uncertainty estimation using deep
  ensembles.
\newblock {\em arXiv preprint arXiv:1612.01474}, 2016.

\bibitem{lane2015deepear}
N.~D. Lane, P.~Georgiev, and L.~Qendro.
\newblock Deepear: robust smartphone audio sensing in unconstrained acoustic
  environments using deep learning.
\newblock In {\em Proceedings of the 2015 ACM International Joint Conference on
  Pervasive and Ubiquitous Computing}, pages 283--294. ACM, 2015.

\bibitem{lim2011investigating}
B.~Y. Lim and A.~K. Dey.
\newblock Investigating intelligibility for uncertain context-aware
  applications.
\newblock In {\em Proceedings of the 13th international conference on
  Ubiquitous computing}, pages 415--424. ACM, 2011.

\bibitem{lipton2016mythos}
Z.~C. Lipton.
\newblock The mythos of model interpretability.
\newblock {\em arXiv preprint arXiv:1606.03490}, 2016.

\bibitem{mannini2013activity}
A.~Mannini, S.~S. Intille, M.~Rosenberger, A.~M. Sabatini, and W.~Haskell.
\newblock Activity recognition using a single accelerometer placed at the wrist
  or ankle.
\newblock {\em Medicine and science in sports and exercise}, 45(11):2193, 2013.

\bibitem{melgarejo2014leveraging}
P.~Melgarejo, X.~Zhang, P.~Ramanathan, and D.~Chu.
\newblock Leveraging directional antenna capabilities for fine-grained gesture
  recognition.
\newblock In {\em Proceedings of the 2014 ACM International Joint Conference on
  Pervasive and Ubiquitous Computing}, pages 541--551. ACM, 2014.

\bibitem{park2011gesture}
T.~Park, J.~Lee, I.~Hwang, C.~Yoo, L.~Nachman, and J.~Song.
\newblock E-gesture: a collaborative architecture for energy-efficient gesture
  recognition with hand-worn sensor and mobile devices.
\newblock In {\em Proceedings of the 9th ACM Conference on Embedded Networked
  Sensor Systems}, pages 260--273. ACM, 2011.

\bibitem{pirkl2012robust}
G.~Pirkl and P.~Lukowicz.
\newblock Robust, low cost indoor positioning using magnetic resonant coupling.
\newblock In {\em Proceedings of the 2012 ACM Conference on Ubiquitous
  Computing}, pages 431--440. ACM, 2012.

\bibitem{pu2013whole}
Q.~Pu, S.~Gupta, S.~Gollakota, and S.~Patel.
\newblock Whole-home gesture recognition using wireless signals.
\newblock In {\em Proceedings of the 19th annual international conference on
  Mobile computing \& networking}, pages 27--38. ACM, 2013.

\bibitem{quinonero2006evaluating}
J.~Quinonero-Candela, C.~E. Rasmussen, F.~Sinz, O.~Bousquet, and
  B.~Sch{\"o}lkopf.
\newblock Evaluating predictive uncertainty challenge.
\newblock In {\em Machine Learning Challenges. Evaluating Predictive
  Uncertainty, Visual Object Classification, and Recognising Tectual
  Entailment}, pages 1--27. Springer, 2006.

\bibitem{radu2016towards}
V.~Radu, N.~D. Lane, S.~Bhattacharya, C.~Mascolo, M.~K. Marina, and F.~Kawsar.
\newblock Towards multimodal deep learning for activity recognition on mobile
  devices.
\newblock In {\em Proceedings of the 2016 ACM International Joint Conference on
  Pervasive and Ubiquitous Computing: Adjunct}, pages 185--188. ACM, 2016.

\bibitem{rasmussen2006Gaussian}
C.~E. Rasmussen.
\newblock Gaussian processes for machine learning.
\newblock 2006.

\bibitem{srivastava2014dropout}
N.~Srivastava, G.~E. Hinton, A.~Krizhevsky, I.~Sutskever, and R.~Salakhutdinov.
\newblock Dropout: a simple way to prevent neural networks from overfitting.
\newblock {\em Journal of Machine Learning Research}, 15(1):1929--1958, 2014.

\bibitem{stisen2015smart}
A.~Stisen, H.~Blunck, S.~Bhattacharya, T.~S. Prentow, M.~B. Kj{\ae}rgaard,
  A.~Dey, T.~Sonne, and M.~M. Jensen.
\newblock Smart devices are different: Assessing and mitigatingmobile sensing
  heterogeneities for activity recognition.
\newblock In {\em Proceedings of the 13th ACM Conference on Embedded Networked
  Sensor Systems}, pages 127--140. ACM, 2015.

\bibitem{toscos2012best}
T.~Toscos, K.~Connelly, and Y.~Rogers.
\newblock Best intentions: health monitoring technology and children.
\newblock In {\em Proceedings of the SIGCHI conference on Human Factors in
  Computing Systems}, pages 1431--1440. ACM, 2012.

\bibitem{wang2016hemaapp}
E.~J. Wang, W.~Li, D.~Hawkins, T.~Gernsheimer, C.~Norby-Slycord, and S.~N.
  Patel.
\newblock Hemaapp: noninvasive blood screening of hemoglobin using smartphone
  cameras.
\newblock In {\em Proceedings of the 2016 ACM International Joint Conference on
  Pervasive and Ubiquitous Computing}, pages 593--604. ACM, 2016.

\bibitem{wang2016simple}
H.~Wang, Y.-H. Kuo, D.~Kifer, and Z.~Li.
\newblock A simple baseline for travel time estimation using large-scale trip
  data.
\newblock In {\em Proceedings of the 24th ACM SIGSPATIAL International
  Conference on Advances in Geographic Information Systems}, page~61. ACM,
  2016.

\bibitem{weppner2016monitoring}
J.~Weppner, B.~Bischke, and P.~Lukowicz.
\newblock Monitoring crowd condition in public spaces by tracking mobile
  consumer devices with wifi interface.
\newblock In {\em Proceedings of the 2016 ACM International Joint Conference on
  Pervasive and Ubiquitous Computing: Adjunct}, pages 1363--1371. ACM, 2016.

\bibitem{yao2016learning}
L.~Yao, F.~Nie, Q.~Z. Sheng, T.~Gu, X.~Li, and S.~Wang.
\newblock Learning from less for better: semi-supervised activity recognition
  via shared structure discovery.
\newblock In {\em Proceedings of the 2016 ACM International Joint Conference on
  Pervasive and Ubiquitous Computing}, pages 13--24. ACM, 2016.

\bibitem{yao2016recursive}
S.~Yao, M.~T. Amin, L.~Su, S.~Hu, S.~Li, S.~Wang, Y.~Zhao, T.~Abdelzaher,
  L.~Kaplan, C.~Aggarwal, et~al.
\newblock Recursive ground truth estimator for social data streams.
\newblock In {\em Information Processing in Sensor Networks (IPSN), 2016 15th
  ACM/IEEE International Conference on}, pages 1--12. IEEE, 2016.

\bibitem{yao2016source}
S.~Yao, S.~Hu, S.~Li, Y.~Zhao, L.~Su, L.~Kaplan, A.~Yener, and T.~Abdelzaher.
\newblock On source dependency models for reliable social sensing: Algorithms
  and fundamental error bounds.
\newblock In {\em Distributed Computing Systems (ICDCS), 2016 IEEE 36th
  International Conference on}, pages 467--476. IEEE, 2016.

\bibitem{yao2017deepsense}
S.~Yao, S.~Hu, Y.~Zhao, A.~Zhang, and T.~Abdelzaher.
\newblock Deepsense: A unified deep learning framework for time-series mobile
  sensing data processing.
\newblock In {\em Proceedings of the 26th International Conference on World
  Wide Web}, pages 351--360. International World Wide Web Conferences Steering
  Committee, 2017.

\bibitem{yao2017deepiot}
S.~Yao, Y.~Zhao, A.~Zhang, L.~Su, and T.~Abdelzaher.
\newblock Deepiot: Compressing deep neural network structures for sensing
  systems with a compressor-critic framework.
\newblock {\em arXiv preprint arXiv:1706.01215}, 2017.

\bibitem{yeo2016watchmi}
H.-S. Yeo, J.~Lee, A.~Bianchi, and A.~Quigley.
\newblock Watchmi: pressure touch, twist and pan gesture input on unmodified
  smartwatches.
\newblock In {\em Proceedings of the 18th International Conference on
  Human-Computer Interaction with Mobile Devices and Services}, pages 394--399.
  ACM, 2016.

\bibitem{zhang2017triovecevent}
C.~Zhang, L.~Liu, D.~Lei, Q.~Yuan, H.~Zhuang, T.~Hanratty, and J.~Han.
\newblock Triovecevent: Embedding-based online local event detection in
  geo-tagged tweet streams.
\newblock In {\em Proceedings of the 23rd ACM SIGKDD International Conference
  on Knowledge Discovery and Data Mining}, pages 595--604. ACM, 2017.

\bibitem{zhang2017regions}
C.~Zhang, K.~Zhang, Q.~Yuan, H.~Peng, Y.~Zheng, T.~Hanratty, S.~Wang, and
  J.~Han.
\newblock Regions, periods, activities: Uncovering urban dynamics via
  cross-modal representation learning.
\newblock In {\em Proceedings of the 26th International Conference on World
  Wide Web}, pages 361--370. International World Wide Web Conferences Steering
  Committee, 2017.

\bibitem{zhang2015eye}
Y.~Zhang, M.~K. Chong, J.~M{\"u}ller, A.~Bulling, and H.~Gellersen.
\newblock Eye tracking for public displays in the wild.
\newblock {\em Personal and Ubiquitous Computing}, 19(5-6):967--981, 2015.

\end{thebibliography}

\end{document}